\PassOptionsToPackage{table,xcdraw,usenames,dvipsnames}{xcolor}
\documentclass{article}

\usepackage{microtype}
\usepackage{graphicx}
\usepackage{subfigure}
\usepackage{booktabs} % for professional tables

\usepackage{hyperref}

\usepackage[accepted]{icml2025}

\usepackage{amsmath}
\usepackage{amssymb}
\usepackage{mathtools}
\usepackage{amsthm}

\usepackage[capitalize,noabbrev]{cleveref}

\theoremstyle{plain}

\theoremstyle{definition}

\theoremstyle{remark}

\usepackage[textsize=tiny]{todonotes}

\usepackage{xspace}
\newcommand{\ourmethod}{{\fontfamily{lmtt}\selectfont \textbf{EvoFlow}}\xspace}
\newcommand{\llmname}[1]{{\fontfamily{pcr}\selectfont {#1}}\xspace}

\usepackage{fp}

\definecolor{ForestGreen}{RGB}{34,139,34}
\definecolor{myyellow}{RGB}{181, 181, 27}
\newcommand{\greencheck}{{\color{ForestGreen}\ding{52}}}

\newcommand{\redcheck}{{\color{red}\ding{55}}}

\usepackage[most]{tcolorbox}
\definecolor{mygrey}{gray}{0.4}

\usepackage[ruled,algo2e]{algorithm2e}
\renewcommand{\algorithmicrequire}{\textbf{Input:}}  % Use Input in the format of Algorithm
\renewcommand{\algorithmicensure}{\textbf{Output:}} % Use Output in the format of Algorithm
\SetKwInOut{Input}{Input}\SetKwInOut{Output}{Output}
\SetKwComment{Comment}{$\triangleright$\ }{}
\definecolor{hengpink}{cmyk}{0, 0.7808, 0.4429, 0.1412}

\usepackage{listings}
\usepackage{multirow}
\usepackage{makecell}
\usepackage{enumerate}
\usepackage{enumitem}
\usepackage{pifont}

\newcommand{\blue}[1]{$_{\color{BlueGreen}\downarrow #1}$}
\newcommand{\red}[1]{$_{\color{RedOrange}\uparrow #1}$}

\usepackage{fixltx2e}

\icmltitlerunning{EvoFlow: Evolving Diverse Agentic Workflows On The Fly}

\begin{document}

\twocolumn[
\icmltitle{EvoFlow: Evolving Diverse Agentic Workflows On The Fly}

% It is OKAY to include author information, even for blind
% submissions: the style file will automatically remove it for you
% unless you've provided the [accepted] option to the icml2025
% package.

% List of affiliations: The first argument should be a (short)
% identifier you will use later to specify author affiliations
% Academic affiliations should list Department, University, City, Region, Country
% Industry affiliations should list Company, City, Region, Country

% You can specify symbols, otherwise they are numbered in order.
% Ideally, you should not use this facility. Affiliations will be numbered
% in order of appearance and this is the preferred way.
\icmlsetsymbol{equal}{*}

\begin{icmlauthorlist}
\icmlauthor{Guibin Zhang}{equal,tongji}
\icmlauthor{Kaijie Chen}{equal,tongji}
\icmlauthor{Guancheng Wan}{whu}
\icmlauthor{Heng Chang}{tsinghua}\\
\icmlauthor{Hong Cheng}{cuhk}
\icmlauthor{Kun Wang}{ntu}
\icmlauthor{Shuyue Hu}{ailab}
%\icmlauthor{}{sch}
\icmlauthor{Lei Bai}{ailab}
%\icmlauthor{}{sch}
%\icmlauthor{}{sch}
\end{icmlauthorlist}

\icmlaffiliation{cuhk}{The Chinese University of Hong Kong}
\icmlaffiliation{tongji}{Tongji University}
\icmlaffiliation{whu}{Wuhan University}
\icmlaffiliation{tsinghua}{Tsinghua University}
\icmlaffiliation{ntu}{Nanyang Technological University}
\icmlaffiliation{ailab}{Shanghai AI Laboratory}

\icmlcorrespondingauthor{Guibin Zhang}{guibinz@outlook.com}

% You may provide any keywords that you
% find helpful for describing your paper; these are used to populate
% the "keywords" metadata in the PDF but will not be shown in the document
\icmlkeywords{Machine Learning, ICML}

\vskip 0.3in
]

% this must go after the closing bracket ] following \twocolumn[ ...

% This command actually creates the footnote in the first column
% listing the affiliations and the copyright notice.
% The command takes one argument, which is text to display at the start of the footnote.
% The \icmlEqualContribution command is standard text for equal contribution.
% Remove it (just {}) if you do not need this facility.

%\printAffiliationsAndNotice{}  % leave blank if no need to mention equal contribution
\printAffiliationsAndNotice{\icmlEqualContribution} % otherwise use the standard text.

\begin{abstract}
The past two years have witnessed the evolution of large language model (LLM)-based multi-agent systems from labor-intensive manual design to partial automation (\textit{e.g.}, prompt engineering, communication topology) and eventually to fully automated design. However, existing agentic automation pipelines often lack LLM heterogeneity and focus on single-objective performance optimization, limiting their potential to combine weaker models for more customized and cost-effective solutions. To address this challenge, we propose \ourmethod, a niching evolutionary algorithm-based framework to automatically search a population of heterogeneous and complexity-adaptive agentic workflows, rather than a single homogeneous, complex workflow. Technically, \ourmethod performs \textit{(1) tag-based retrieval} to extract parent workflows from an agentic population, evolves new workflows through \textit{(2) crossover} and \textit{(3) mutation}, and employs \textit{(4) niching-based selection} to maintain population diversity and quality. Extensive evaluations across seven benchmarks demonstrate that \ourmethod is:  
\textbf{(I) diverse}, evolving a population of workflows ranging from simple I/O tasks to complex multi-turn interactions;  
\textbf{(II) high-performing}, outperforming previous handcrafted and automated workflows by $1.23\%\sim29.86\%$;  
\textbf{(III) economical}, surpassing powerful \llmname{o1-preview} at $12.4\%$ of its inference cost using weaker open-source models. The code will be available at \url{https://github.com/bingreeky/EvoFlow}.
\end{abstract}

\vspace{-2em}
\section{Introduction}
\vspace{-0.4em}
Large Language Model (LLM)-based agents \citep{autogpt,babyagi,agentgpt} have exhibited remarkable capabilities across a wide spectrum of tasks, including question answering~\cite{zhu2024autotqa}, data analysis~\citep{hong2024datainterpreter,li2024autokaggle}, decision-making~\citep{song2023llmplanner}, code generation~\citep{reflexion}, video gaming \citep{voyager}, and autonomous driving \citep{jin2023surrealdriver}, among others. Recent advancements further highlight that integrating single agents into \textit{agentic workflows}, \textit{i.e.}, structured sequences of LLM-based agent interactions, can surpass the cognitive and functional limitations of individual agents \citep{arXiv2023_MultiAgent-Debate, arXiv2023_MultiAgent-Debate_2, multi-persona, blender, reflexion, PHPrompting, autogen, zhang2024cut}, thereby exhibiting human-esque collaborative intelligence in multi-agent systems~\citep{zhang2023exploring}.

\begin{figure}[t]
\centering
\includegraphics[width=1\columnwidth]{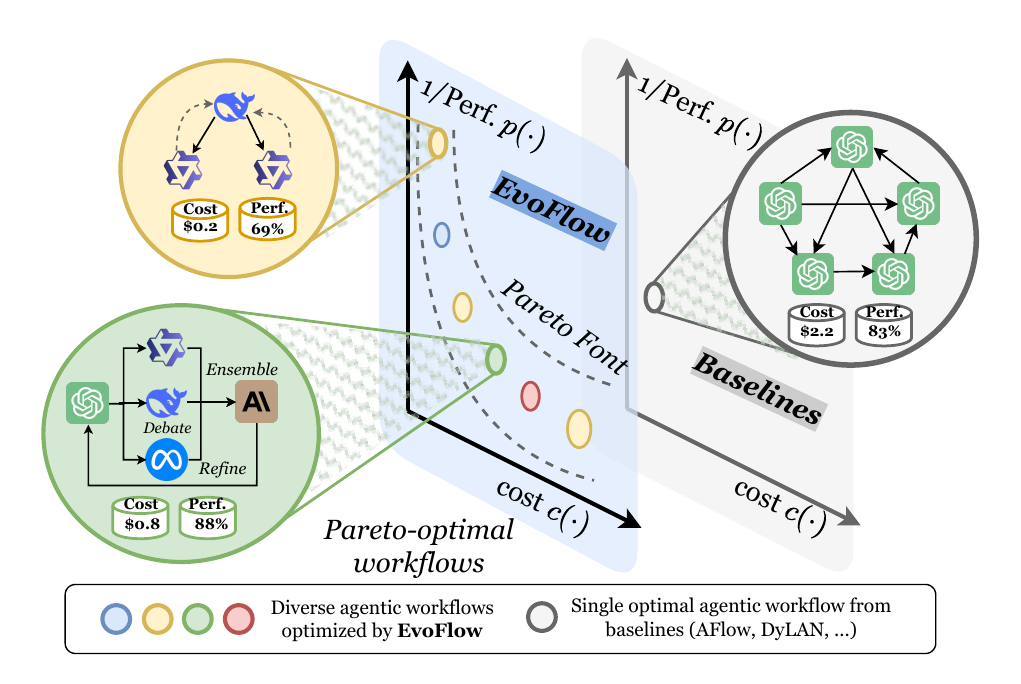}
\vspace{-1cm}
\caption{Paradigm comparison. Baseline methods seek a ``one-size-fits-all'' complex homogenoues workflow, while \ourmethod optimizes a Pareto set of diverse, heterogenous workflows. 
% The red and blue lines represent the ROC-AUC  (left $y$-axis) and the spectral preservation ratio\footnotemark of the sparse subgraph (right $y$-axis), respectively. The horizontal red dashed line indicates the original performance. 
%It is observed that with increasing sparsity, both sparsifiers result in decreased performance and spectral preservation, yet the decline with Local Degree is more gradual.
}
\vspace{-1.8em}
\label{fig:intro}
\end{figure}

The progress of machine learning consistently reveals a recurring pattern: manually crafted artifacts are often replaced by learnable and optimizable ones~\citep{tyson2022automation,clune2019ai}. For natural language processing (NLP), this is exemplified by the replacement of hand-designed representations like Bag-of-Words~\citep{mikolov2013efficient} with learned embeddings like GloVe~\citep{pennington2014glove} and BERT~\citep{devlin2018bert}. Similarly, agentic systems have experienced a rapid transition from manual to automated over the past two years: Early attempts, \textit{e.g.}, CAMEL~\citep{NeurIPS2023camel}, AutoGen~\citep{autogen}, and MetaGPT~\citep{meta-gpt}, relied heavily on manual configurations, while follow-up works significantly reduced dependence on human intervention, like DsPy~\citep{khattab2023dspy} automating prompt optimization, GPTSwarm~\citep{zhuge2024gptswarm} optimizing inter-agent communication, and EvoAgent~\citep{yuan2024evoagent} self-evolving agent profiling. More recent efforts~\citep{hu2024adas,shang2024agentsquare,zhang2024aflow}, have demonstrated that these automated pipelines enable the development of surprisingly creative and powerful agentic workflows, marking significant progress toward fully autonomous agentic AI. 
Despite their success, existing automation pipelines often lack \textit{diversity} in both LLM heterogeneity and complexity scheduling:

\vspace{-1.em}
\begin{itemize}[leftmargin=*,itemsep=-0.1em]
    \item[\ding{224}] \textbf{\textit{Lack of LLM heterogeneity.}}  Mainstream multi-agent workflows are often \textit{homogeneous}, relying on a single, expensive LLM like GPT-3.5/GPT-4o to instantiate all agents~\cite{zhang2024aflow,arXiv2023_AutoAgents}. However, as highlighted by studies on LLM routing~\cite{chen2023frugalgpt,hu2024routerbench}, the capabilities of different LLMs are often \textit{complementary} rather than forming a strict superset relationship. In certain scenarios, smaller LLMs can perform tasks comparably or even outperform their larger counterparts at a significantly lower token cost. Against this backdrop, we advocate for agentic workflows to be \textit{heterogeneous}, incorporating a diverse ensemble of LLMs with varying sizes, capabilities, and sources.
    %the prevalent homogeneous reliance on expensive models like GPT-3.5/GPT-4 appears highly questionable. 
    % \vspace{-0.3em}
    \item[\ding{224}] \textbf{\textit{Lack of complexity diversity.}}  Existing automated agentic workflows often prioritize single-objective optimization, focusing on performance or solution quality. This typically results in a singular, complex workflow incorporating elements like CoT ensembles and multi-turn discussions~\citep{hu2024adas,shang2024agentsquare}. However, real-world user queries vary significantly in difficulty, as exemplified by the MMLU benchmark~\cite{mmlu}, which spans tasks from elementary to graduate-level. While complex workflows are essential for the latter, simpler queries can be efficiently addressed by single-agent I/O~\citep{feng2024graphrouter}. Thus, we advocate for optimizing a diverse set of workflows, tailoring simple workflows to straightforward tasks and reserving complex ones for more intricate challenges.
\end{itemize}
\vspace{-0.8em}

The above considerations and observations raise critical questions regarding the current paradigm of agentic system design: \textit{How can we automatically optimize a set of heterogeneous, complexity-adaptive agentic workflows that provide diverse solutions for varied queries?}

To this end, we propose \ourmethod, a niching evolutionary algorithm (EA)-based framework for automatically searching a population of heterogeneous, complexity-adaptive agentic workflows, rather than a single, homogeneous, complex workflow. Technically, \ourmethod innovatively frames the agentic search as a multi-objective optimization problem, considering both cost and performance, ultimately generating a Pareto-optimal set of workflows balancing these factors. \ourmethod uses \textbf{operator nodes}, \textit{i.e.}, a set of LLM-agent \textbf{invoking nodes}, as the fundamental units of its search space. The workflow population is initialized by selecting and combining multiple operator nodes. 
Afterward, \ourmethod continuously evolves by processing incoming queries. It \textbf{\ding{182} tag-based retrieves} relevant workflows as parents, performs \textbf{\ding{183} crossover} to generate offspring workflows, applies \textbf{\ding{184} mutation} with extensive mutation functions, \textit{i.e.}, LLM/prompt/operator mutation, to evolve the offsprings.
 Finally, \textbf{\ding{185} niching selection} is leveraged to maintain diversity and quality in the population. During inference, \ourmethod autonomously retrieves domain-relevant and complexity-adapted agentic systems from the well-optimized population, to swiftly and efficiently address user queries.

We conduct comprehensive evaluations on six widely adopted benchmarks.  \textbf{In heterogeneous settings}, \ourmethod surpasses powerful \llmname{o1-preview} at $12.4\%$ of its inference cost by utilizing weaker open-source models (\textit{e.g.}, \llmname{LLaMa-3.1-70b} and \llmname{QWen-2.5-72b}); \textbf{In homogeneous settings}, \ourmethod outperforms state-of-the-art (SOTA) agentic workflows by an average of $1.23\%\sim29.86\%$ in performance. More importantly, \ourmethod is \textbf{highly economical}, with a training cost of only one-third of SOTA baseline AFlow ($\$0.45$ vs $\$1.23$) and an inference cost of merely one-fifth ($\$0.51$ vs $\$2.62$), while surpassing AFlow by $5.91\%\uparrow$ on the MATH Benchmark.

Briefly put, our contributions can be summarized as:

\vspace{-1.em}
\begin{itemize}[leftmargin=*,itemsep=-0.1em]
\item \textbf{Paradigm Transformation.} We \textit{for the first time} explicitly formulate agentic workflow automation as a cost-performance multi-objective optimization problem, highlighting \textit{LLM heterogeneity} and \textit{complexity diversity} as key features for the development of multi-agent systems.
\item \textbf{Practical Solution.} We propose a niching evolutionary algorithm-based framework, \ourmethod, which autonomously evolves a population of heterogeneous and complexity-diverse agentic workflows across various task domains with minimal human intervention.  
    \item \textbf{Emperical Evaluation.} Extensive experiments on seven benchmarks show that \ourmethod is \textbf{(I) diverse}, evolving a workflow population ranging from simple I/O to complex multi-turn interactions; \textbf{(II) high-performing}, surpassing previous handcrafted and automated workflows by $1.23\%\sim29.86\%$; \textbf{(III) economical}, surpassing powerful \llmname{o1-preview} with weaker open-source models.
\end{itemize}
\vspace{-0.5em}

\vspace{-0.5em}
\section{Related Work}
\vspace{-0.5em}
\paragraph{LLM-based Autonomous Agents}
% \vspace{-0.5em}
%With the advent of advanced large language models (LLMs)~\citep{openai2023gpt4,team2023gemini}, significant efforts have been made to develop autonomous agents~\citep{autogpt,babyagi} by equipping LLMs with high-level features, such as persona, tool, planning and memory~\citep{shen2024hugginggpt,zhu2024knowagent,zhong2024memorybank}. 
Building on the success of single agent~\citep{shen2024hugginggpt,zhu2024knowagent,zhong2024memorybank}, studies have shown that interaction among multiple LLM-based agents can substantially enhance individual model capabilities~\citep{FCS2024_Survey-Agent}, as seen in several early frameworks, including CAMEL~\citep{NeurIPS2023camel}, AutoGen~\citep{autogen}, BabyAGI~\citep{babyagi}, and LLM-Debate~\citep{arXiv2023_MultiAgent-Debate}. However, these initial approaches heavily depended on manually crafted designs, which constrained the adaptability and flexibility of agents in addressing unforeseen challenges~\citep{he2023lego,chen2023agentverse}. Consequently, the push toward automating agentic workflows has gained momentum.

\vspace{-0.7em}
\paragraph{Automated Agentic Workflows}
% \vspace{-0.5em}
Efforts to automate agentic workflows can be broadly categorized into the following types: \textbf{(1) Prompt Optimization}, exemplified by PromptBreeder~\citep{fernando2023promptbreeder} and DsPy~\citep{khattab2023dspy}; \textbf{(2) Inter-agent Topology}, which focuses on orchestrating interactions among agents, such as GPTSwarm~\citep{zhuge2024gptswarm}, DyLAN~\citep{arXiv2023_Dynamic-LLM-Agent}, EvoMAC~\citep{hu2024evomac}, and G-Designer~\citep{zhang2024gdesigner}; \textbf{(3) Agent Persona/Profile}, represented by AgentVerse~\citep{chen2023agentverse} and EvoAgent~\citep{yuan2024evoagent}. More recently, \citet{hu2024adas} formalized the concept of \textit{Automated Design of Agentic Systems}, with subsequent advancements by AgentSquare~\citep{shang2024agentsquare} and AFlow~\citep{zhang2024aflow}. However, these automation pipelines are predominantly \textit{homogeneous}, \textit{i.e.}, utilizing a single-source LLM, and lack the integration of \textit{heterogeneous} LLM agents of varying sizes and sources. Additionally, they typically produce a fixed workflow~\citep{yuan2024evoagent,zhuge2024gptswarm,zhang2024aflow}, which cannot dynamically allocate resources when confronted with tasks/queries of different levels and complexities.

\vspace{-1em}
\begin{table}[ht!]
\caption{Comparison among different automation techniques. }
\label{tab:intro_compare}
\scriptsize
\setlength\tabcolsep{0.8pt}
\resizebox{\columnwidth}{!}{\begin{tabular}{l|ccc|cc}
\hline
 {\textbf{Method}} & \makecell[c]{\textbf{Prompt}\\ \textbf{Optimize}} & \makecell[c]{\textbf{Agent}\\ \textbf{Topology}} & \makecell[c]{\textbf{Agent}\\ \textbf{Profile}} & \makecell[c]{\textbf{LLM}\\ \textbf{Backbone}} &  \makecell[c]{\textbf{Complexity}\\ \textbf{Adaptivity}}  \\ \hline 
 {AgentVerse} & \redcheck & \redcheck & \greencheck & \redcheck & \redcheck  \\
{GPTSwarm}  & \redcheck & \greencheck & \redcheck & \redcheck & \redcheck \\
{EvoMAC} & \greencheck & \greencheck & \redcheck & \redcheck & \redcheck  \\
{EvoAgent} & \greencheck & \redcheck & \greencheck & \redcheck & \redcheck  \\
{EvoPrompt} & \greencheck & \redcheck & \redcheck & \redcheck & \redcheck  \\
% {G-Designer} & \redcheck & \greencheck & \redcheck & \redcheck  & \greencheck \\
{ADAS} & \greencheck & \redcheck & \greencheck & \redcheck & \redcheck \\
{AFlow} &  \greencheck & \greencheck & \greencheck & \redcheck & \redcheck \\
{AgentSquare} & \redcheck & \greencheck & \greencheck & \redcheck & \redcheck \\
\hline
\textbf{\ourmethod} & \greencheck & \greencheck & \greencheck & \greencheck & \greencheck  \\
\hline
 % \hline %\hline 
 % \multicolumn{5}{l}{
 %  \tiny$\S$ PRC: Prune Rate Control
 % }
\end{tabular}}
% \vspace{-1.8em}

\vspace{-1.4em}
\end{table}

\vspace{-0.4em}
\paragraph{Evolutionary Algorithm}
% \vspace{-0.4em}
Evolutionary algorithms (EAs) are no new to agentic AI~\citep{cetnarowicz1996application,li2016multi,liu2020mapper}. In the era of LLM-based agents, researchers have explored the interplay between EA and LLM agents, including prompt engineering~\citep{xu2022gps,shi2024red}, code generation~\citep{romera2024mathematical}, project planning~\citep{tao2023program} and inference time scaling~\citep{lee2025evolving-deepmind}. EvoAgent~\citep{yuan2024evoagent} and EvoPrompt~\citep{guo2023evoprompt} employ simple genetic algorithms to optimize agent profiles and prompts, whose, however, level of automation is highly constrained, focusing solely on single-agent prompt optimization and failing to evolve at the workflow level, as illustrated in \Cref{tab:intro_compare}.

% 在AutoML中，evolutionary algorithm (EA)是被广泛采用的技术手段，在hyperparameter searching以及neural architecture search (NAS)等领域。When it comes to agentic systems，同样存在一些EA-based 早期工作，如
                                                                
\vspace{-0.9em}
\section{Preliminary}
\vspace{-0.5em}
In this section, we formally define the search space of \ourmethod and the objective of workflow optimization.  

\begin{figure}[!h]
\centering
\includegraphics[width=1.0\linewidth]{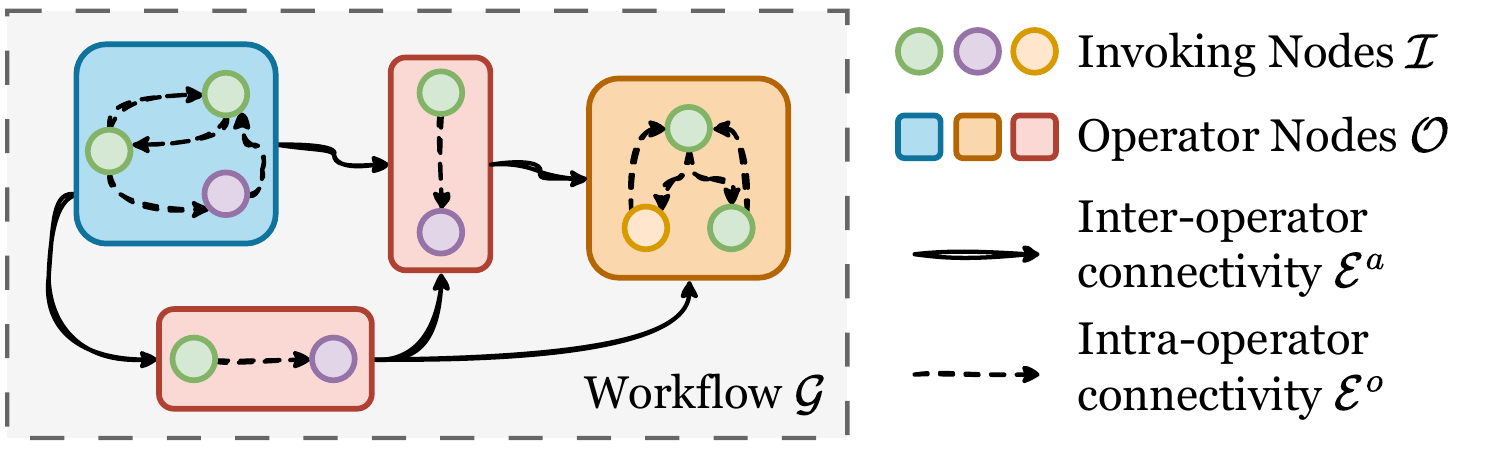}
\vspace{-0.8cm}
\caption{The visualization of notations in \ourmethod.
}
\vspace{-1.5em}
\label{fig:notations}
\end{figure}

\textbf{Search Space.} 
%The rapid advancements in Large Language Models (LLMs) have resulted in an increasingly diverse range of \textit{heterogeneous} models, varying in size, functionality, and computational requirements, which can be utilized to construct agentic workflows. 
The search space of \ourmethod is defined hierarchically, with the basic unit being the (LLM-)invoking node. These are further assembled into (composite) operator nodes, which are then combined to form the complete workflow $\mathcal{G}$, as visualized in \Cref{fig:notations}. Each invoking node ${I}_i$ is defined as follows:
\begin{equation}
{I}_i = (M_i, P_i, \tau_i), \; P_i \in \mathcal{P}, \tau_i\in {[0,1]},
\end{equation}
where $P_i$ represents the associated prompt, with $\mathcal{P}$ denoting the feasible prompt space, and $\tau_i$ is the temperature parameter. $M_i = (|M_i|, C_i, L_i)$ represents an LLM instance from the feasible model pool $\mathcal{M} = \{M_1, \cdots, M_{|\mathcal{M}|}\}$, characterized by its model size $|M_i|$, token cost $C_i$, and inference delay $L_i$. Thus, the feasible space for invoking nodes is given by $\mathcal{I} = \mathcal{M} \times \mathcal{P} \times \mathbb{R}_{[0,1]}$. Notably, prior agentic automation pipelines have generally excluded $\mathcal{M}$ from their search space~\cite{zhang2024aflow,yuan2024evoagent,zhuge2024gptswarm}, often preselecting a single LLM $M$ to instantiate all agents, which constrains the development of more diverse and capability-rich agentic systems.
Building upon the invoking nodes, the operator node ${O}_j$ is represented by:
\begin{equation}
{O}_j = (\mathcal{I}^o_j, \mathcal{E}^o_j),\mathcal{I}^o_j = \{I_1, \dots, I_n\},\mathcal{E}^o_j \subseteq \mathcal{I}^o_j \times\mathcal{I}^o_j,
\end{equation}
where $\mathcal{I}^o_j$ is a subset of invoking nodes, and $\mathcal{E}^o_j$ signifies the connectivity relationship among invoking nodes.
The overall agentic workflow $\mathcal{G}$ is defined as:
\begin{equation}
\small
\begin{aligned}
 \mathcal{G} & = (\mathcal{O}^S, \mathcal{E}^a), \mathcal{O}^S = \{O_1, \dots, O_m\}, \mathcal{E}^a \subseteq \mathcal{O}^S \times \mathcal{O}^S,\\
 & =(\mathcal{I}^S, \mathcal{E}^o),  \mathcal{I}^S = \bigcup_{j=1}^m \mathcal{I}^o_j, \mathcal{E}^o = \bigcup_{j=1}^m \mathcal{E}^o_j \cup \mathcal{E}^a,
 \end{aligned}
\end{equation}
where $\mathcal{O}^S\subseteq\mathcal{O}$, $\mathcal{I}^S\subseteq\mathcal{I}$, $m$ denotes the number of operator nodes in $\mathcal{G}$, \(\mathcal{E}^a\)/\(\mathcal{E}^o\) denote intra/inter-operator connections.

% \begin{equation}
% \mathcal{G}_k=(\mathcal{L}^a_k, \mathcal{E}^a_k)=(\mathcal{O}_k,\mathcal{E}^o_k)
% \end{equation}

% $
% \mathcal{G}=(\mathcal{O},\mathcal{E}_g)$, where $\mathcal{O}=\{O_1,\cdots,O_m\}$ is the operator node set and $\mathcal{E}_g\subseteq  \mathcal{O}\times \mathcal{O}$ denotes the operator interactions.

% Consider an available LLM pool denoted as $\mathcal{M} = \{M_1, \cdots, M_{|\mathcal{M}|}\}$, where each LLM backbone $M_i=(|M_i|, C_i,L_i)$ is characterized by its model size $|M_i|$, token cost $C_i$ and inference delay $L_i$. Notably, prior agentic automation pipelines have generally excluded $\mathcal{M}$ from their search space, often preselecting a single LLM to instantiate all agents, which constrains the development of more diverse and capability-rich agentic systems. 我们进一步介绍the invoking node，which serve as 构成agentic workflow的最基本元素$v_i=$

\begin{figure*}[!t]
\centering
\includegraphics[width=1.0\linewidth]{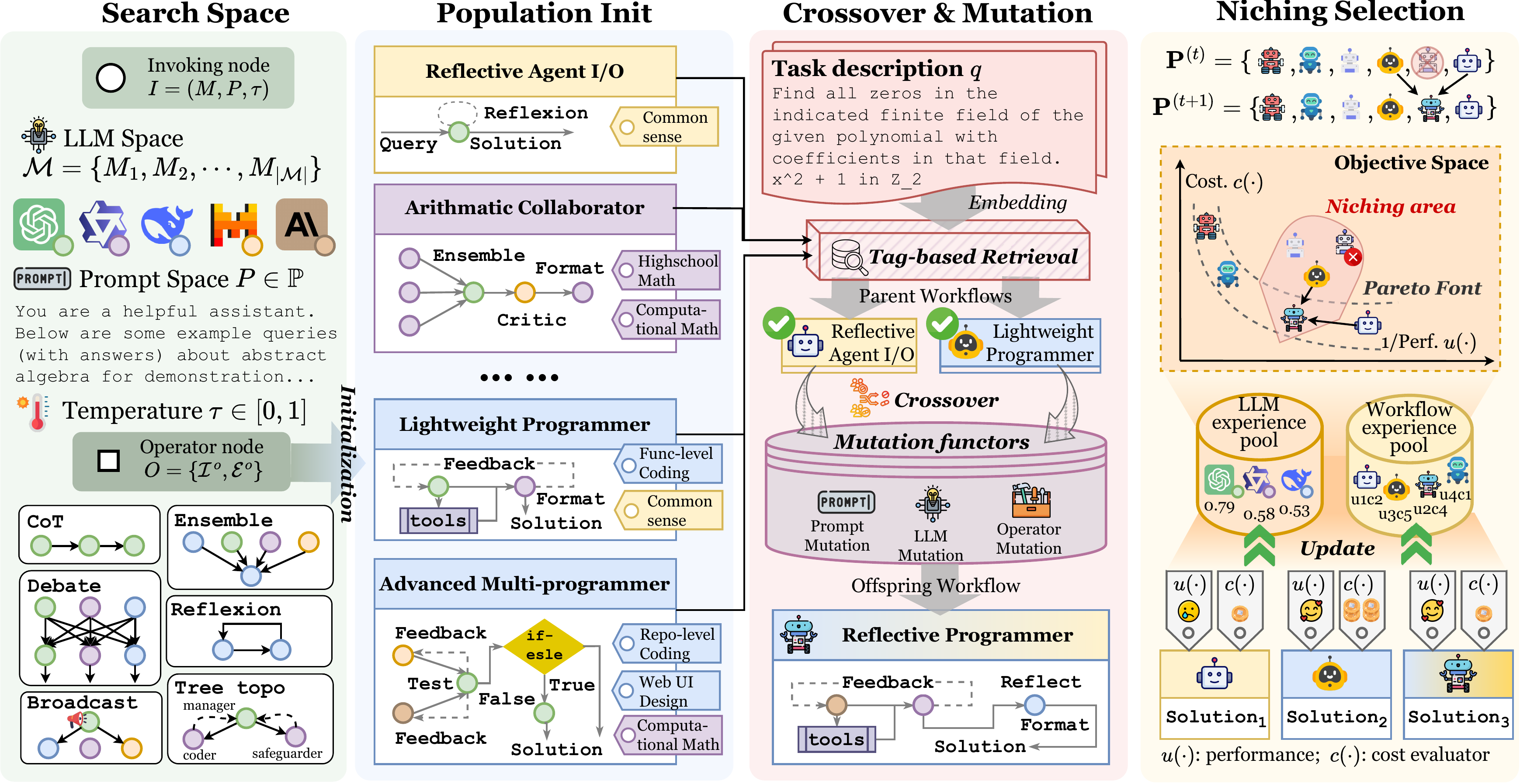}
\vspace{-0.8cm}
\caption{The overall framework of \ourmethod. The fundamental unit is the invoking nodes, which collectively form the operator node. \ourmethod initializes the population by combining multiple operator nodes into a workflow (individual), followed by tag-based retrieval and crossover \& mutation to generate novel offspring workflows. The population is updated via niching-based selection. 
}
\vspace{-1.2em}
\label{fig:framework}
\end{figure*}

\textbf{Problem Formulation} We first present the optimization objective for traditional agentic automation methods. Given a task domain \(T\) and an performance evaluator function \(u(\cdot)\), the objective function is defined as:
\begin{equation}\label{eq:original_objective}
\small
\mathcal{G}^* = \underset{\mathcal{G} \in \mathcal{H}(\mathcal{I}, \mathcal{E})}{\arg\max} \; u(\mathcal{G},T)=\underset{{\scriptsize \mathcal{I}^S \subseteq \mathcal{I}, \mathcal{E}\subseteq\mathcal{I}^S\times \mathcal{I}^S  }}{\arg\max} \!\! u\left((\mathcal{I}^S, \mathcal{E}),T\right),
\end{equation}
where \(\mathcal{I}\) represents the feasible space of invoking nodes, and \(\mathcal{H}(\mathcal{I}, \mathcal{E}^o)\) denotes the invoking node-based search space for \(\mathcal{G}\). As shown in \Cref{eq:original_objective}, existing methods~\cite{zhang2024aflow, shang2024agentsquare, zhuge2024gptswarm} typically perform single-objective optimization. In contrast, the optimization objective of \ourmethod is multi-objective:
\begin{equation}\label{eq:obj_1}
\mathcal{G}^\star = \underset{\mathcal{G} \in \mathcal{H}(\mathcal{I}, \mathcal{E}^o)}{\arg\max} \; \left[ u(\mathcal{G}, T), -c(\mathcal{G}, T) \right]^\top,
\end{equation}
where \(c(\cdot)\) evaluates the system cost, and \(\mathcal{G}^\star\) represents the Pareto optimal set balancing cost and performance, which includes a set of non-dominated agentic workflows that are well-distributed and located near the Pareto front (PF) in the objective space. Detailed explanations are in \Cref{app:obj}. 

However, the optimization in \Cref{eq:obj_1} is currently based on invoking nodes, which results in an excessively large search space and does not explicitly account for many already-existing high-performing composite structures, such as CoT~\citep{cot}, ToT~\cite{tot}, and Multi-agent Debate~\cite{arXiv2023_MultiAgent-Debate}. Therefore, we reformulate the search space to be operator node-based:
\begin{equation}
\small
\begin{aligned}
\mathcal{G}^\star & = \underset{\mathcal{G} \in \mathcal{H}(\mathcal{O}, \mathcal{E}^a)}{\arg\max} \; \left[ u(\mathcal{G}, T), -c(\mathcal{G}, T) \right]^\top,\\
& = \!\!\!\underset{\mathcal{O}^S\subseteq\mathcal{O}, \mathcal{E}^a\subseteq\mathcal{O}^S\times \mathcal{O}^S }{\arg\max}{\!\!\!\!\!\!\! [u((\mathcal{O}^S,\mathcal{E}^a), T), -c((\mathcal{O}^S,\mathcal{E}^a), T)]^\top},
\end{aligned}
\end{equation}
Where \(\mathcal{O}\) represents the feasible space for operator nodes.
%which can incorporate existing structures, \textit{e.g.}, ReAct~\cite{yao2023react} and Reflexion~\cite{reflexion}. %The following sections will detail how \ourmethod efficiently conducts multi-objective agentic workflow search using operator nodes as the fundamental unit.

% generally evaluate \(u(\cdot)\) based on performance, aiming to optimize and discover the best workflow \(\mathcal{G}^*\) for a given task domain, as evident in state-of-the-art works 于此不同的是，\ourmethod formulate a 

% We argue that tasks within the same domain can vary significantly in difficulty. For instance, the MMLU~\citep{mmlu} benchmark includes problems spanning from elementary school to Ph.D. level, with the former requiring simple I/O from a single agent and the latter necessitating extensive debate and collaboration among multiple agents. Thus, we formulate a multi-objective optimization problem rather than single-objective:

\vspace{-0.4em}
\section{Methodology}
\vspace{-0.5em}
% \subsection{Overview}
% \vspace{-0.5em}
As shown in \Cref{fig:framework}, \ourmethod initializes a set of workflows with varying complexities and domain specializations, each tagged with its domain expertise (\Cref{sec:population-init}). As new user queries arrive, \ourmethod performs tag-based retrieval to select the most relevant (workflow) individuals as parents and generates offspring workflows through crossover and mutation (\Cref{sec:offspring}). Upon receiving environmental feedback, the framework evaluates the new workflows in the multi-objective space, and conducts environmental selection to maintain population diversity and efficiency (\Cref{sec:niching}).
    
\vspace{-0.5em}
\subsection{Population Initialization}\label{sec:population-init}
\vspace{-0.5em}
\ourmethod initially populates the feasible space of operator nodes with a basket of powerful single-/multi-agent structures, including CoT~\cite{cot}, Ensemble~\citep{blender}, Self-Reflexion~\cite{reflexion}, Multi-agent Debate~\citep{arXiv2023_MultiAgent-Debate}, \textit{etc}. Detailed formalizations are in \Cref{app:operator}. This transforms $\mathcal{O}$ into a finite set:  
\begin{equation}  
\mathcal{O}^{(0)}=\{O_\text{CoT}, O_\text{Reflexion}, \cdots, O_\text{Debate}\}.  
\end{equation}  
Importantly, thought $\mathcal{O}^{(0)}$ is initialized as a finite set, this does not constrain \ourmethod's potential to explore a broader search space. This is because (1) practitioners can easily customize operator templates as needed, and (2) \ourmethod's crossover and mutation can generate novel operators. 

To initialize the workflow population $\mathbf{P}^{(0)}=\{\mathcal{G}_1, \mathcal{G}_2, \cdots, \mathcal{G}_N\}$ (where the population size is set to $N$), we generate $N$ workflows as follows:  
\begin{equation}  
\begin{gathered}  
\label{eq:init-population}
\mathcal{G}_k \leftarrow \{\mathcal{O}_k, \mathcal{E}^a_k\},  \mathcal{E}^a_k \subseteq \mathcal{O}_k \times \mathcal{O}_k, \\  
\mathcal{O}_k = \{O_i(\mathcal{M}^S_{k,i}, \mathcal{P}_{k,i}^S, \tau_k)\}_{i=1}^m \subseteq \mathcal{O}^{(0)}, \\  
i \sim \operatorname{Uniform}(1, |\mathcal{O}^{(0)}|),  \mathcal{M}_{k,1}^S \subseteq \mathcal{M}, \mathcal{P}_{k,i}^S \subseteq \mathcal{P},  
\end{gathered}  
\end{equation}  
where $m$ denotes the number of operators in workflow $\mathcal{G}_k$, $O_i$ represents the operator template randomly selected from $\mathcal{O}^{(0)}$, and it is instantiated with LLMs $\mathcal{M}_{k,i}^S$ (sampled from LLM pool $\mathcal{M}$) and prompts $\mathcal{P}_{k,i}^S$. Upon creating a workflow individual, we assign it multiple \textit{utility indicator} tags, which suggest the task domains where it might excel. These tags are conducive to a rapid match between user queries and relevant workflows, generated as follows:
%The motivation behind this is straightforward: when a new query arrives, users may struggle to identify the most suitable workflow, as it is typically represented by code or flowcharts~\citep{xue2024genagent}, while domain-specific tags facilitate rapid matching. The generation process is as follows:  
\begin{equation}  \label{eq:tag}
\{\varkappa^k_1, \cdots, \varkappa^k_\kappa\} \leftarrow f_\text{tag}(\mathcal{G}_k),  
\end{equation}  
where $\varkappa_i$ represents the $i$-th tag of $\mathcal{G}_k$, with a total of $\kappa$ tags per workflow, and $f_\text{tag}$ is a LLM-powered tag generation function (see  \Cref{app:prompt_tag}).  Thus, we have initialized the workflow population \(\mathbf{P}^{(0)}\) prior to iterative evolution.

% \vspace{-0.5em}
% \subsection{Tag-based Retrieval}
% \vspace{-0.5em}

\vspace{-0.5em}
\subsection{Retrieval, Crossover, and Mutation}\label{sec:offspring}
\vspace{-0.5em}

Upon initializing  $\mathbf{P}^{(0)}$, we seek to evolve it driven by incoming task queries. Rather than activating the entire population for each query, we select a subset of workflows that are most relevant and complexity-adapted to it, which prevents the population from converging solely toward high-complexity, single-domain evolution.
Specifically, for the \(t\)-th query \(q_t\), we first select the \(K\) most relevant workflow individuals from the population \(\mathbf{P}^{(t)}\), based on utility indicator tags:  
\begin{equation} \label{eq:select-parent}
\begin{aligned}  
\{\mathcal{G}_{t1}, \cdots, \mathcal{G}_{tK}\} &= \operatorname{TopK}\left(\mathcal{S}(\{\mathcal{G}_i\}_{i=1}^N \;|\; q_t), K\right), \\  
\mathcal{S}(\mathcal{G}_i \;|\; q_t) & = \sum_{j=1}^\kappa \frac{\mathbf{v}(\varkappa_{i,j}) \cdot \mathbf{v}(q_t)}{\|\mathbf{v}(\varkappa_{i,j})\| \|\mathbf{v}(q_t)\|},  
\end{aligned}  
\end{equation}  
where \(\varkappa_{i,j}\) is the \(j\)-th tag of \(\mathcal{G}_i\), $\operatorname{TopK}(\cdot,K)$ is a selection function
that outputs elements with $K$ largest values, and \(\mathbf{v}(\cdot)\) maps queries/tags to fixed-length embeddings using lightweight models such as SentenceBERT~\citep{reimers2019sentence} or MiniLM~\citep{wang2020minilm}. The similarity score \(\mathcal{S}(\mathcal{G}_i \mid q_t)\) is computed based on the cosine similarity between the tag/query embeddings.  
After identifying the \(K\) most relevant workflows, these are treated as \textit{parents} to generate \textit{offspring} workflows with \textit{crossover function}:
\begin{equation}
\label{eq:crossover}
\mathcal{G}_\circ^{(t)}\leftarrow \operatorname{Crossover}(\mathcal{G}_{t1}, \cdots, \mathcal{G}_{tK}),
\end{equation}
where \(\mathcal{G}_\circ^{(t)}\) represents the generated workflow, and the \(\operatorname{Crossover}(\cdot)\) function is LLM-facilitated (see prompts in \Cref{app:prompt_generate}).  
To further enhance the diversity of the population and foster the evolution of novel agentic architectures, we apply a suite of \textit{mutation} functions to refine the sketched offspring, as described below:  

\textbf{LLM Mutation $\mu^{l}(\cdot)$} replaces the LLM backbone of an invoking node within an existing workflow. This mutation can be beneficial in scenarios where an agent in the workflow is underperforming, such as when a small 7b agent fails to handle complex subtask decomposition and needs to be replaced by a larger one, or when a simpler 72b model suffices to accomplish a task as a 405b model does. The mutation process empowers \ourmethod to evolve into powerful heterogeneous workflows, formalized as follows:  
\begin{equation}
\label{eq:llm-mutation}
\begin{gathered}
\Tilde{\mathcal{G}} = \mu^{l}(\mathcal{G},\mathcal{R}) = (\mathcal{I}',\mathcal{E}),\\
\mathcal{I}' = \{(M_i', P_i, \tau_i) \mid I_i  \in \mathcal{I}, M_i' = \mathcal{R}(M_i\mid \mathcal{P}_\text{LLM})\},
\end{gathered}
\end{equation}  
where \(\Tilde{\mathcal{G}}\) represents the mutated individual, and  \(\mathcal{R}^l(\cdot)\) is an LLM-powered process that determines whether the LLM in \(I_i\) should be changed based on the LLM performance history pool \(\mathcal{P}_\text{LLM}\). Details of \(\mathcal{P}_\text{LLM}\) and \(\mathcal{R}^l(\cdot)\) are provided in \Cref{app:llm_pool} and \Cref{app:change_llm}, respectively.

\textbf{Prompt Mutation $\mu^{p}(\cdot)$} involves modifying the prompts of invoking nodes, such as incorporating few-shot examples or providing clearer task instructions, as follows:  
\begin{equation}
\label{eq:prompt-mutation}
\begin{gathered}
\small
\Tilde{\mathcal{G}} = \mu^{p}(\mathcal{G},\mathcal{R}^p) = (\mathcal{I}',\mathcal{E}),\\
\mathcal{I}' = \{(M_i, P_i', \tau_i) \mid I_i  \in \mathcal{I}, P_i' = \mathcal{R}^p(P_i\mid \mathcal{P}_\text{wf})\},
\end{gathered}
\end{equation}
where $\mathcal{P}_\text{wf}$ is the experience history of the workflow population, the configuration of which is placed at \Cref{app:workflow_pool}. $\mathcal{R}^p(\cdot)$ is also LLM-powered (see \Cref{app:change_prompt}).

\textbf{Operator Mutation $\mu^{o}(\cdot)$} refers to the modification of the operators or their topological connections ($\mathcal{E}^a\cup\mathcal{E}^o$). Practical scenarios include removing redundant Reflexion operators or adding a ``format'' operator within the workflow to enhance the formatting accuracy of code generation. The process is described as:  
\begin{equation}
\label{eq:operator-mutation}
\begin{gathered}
\small
\Tilde{\mathcal{G}_k} = \mu^{o}(\mathcal{G}_k,\mathcal{R}^o)  = (\mathcal{O}'_k, \mathcal{E}_k^{a'}), \\
\mathcal{O}'_k = \left(\mathcal{O}_k \setminus \mathcal{O}^{del}\right)\cup \mathcal{O}^{add}, \\\mathcal{E}_k^{a'}=\mathcal{R}^o(\mathcal{O}'_k\mid \mathcal{P}_\text{wf})\subseteq \mathcal{O}'_k\times \mathcal{O}'_k,
\end{gathered}
\end{equation}
where $\mathcal{O}^{del}$ and $\mathcal{O}^{add}$ are deleted and added operators by $\mathcal{R}^o$, and $\mathcal{E}_k^{a'}$ denotes the modified topological operator connections. The prompts for $\mathcal{R}^o$ is in \Cref{app:change_topo}.

The mutated offspring is denoted as \({\mathcal{G}_\circledcirc^{(t)}}\). With new individuals introduced, the critical challenge is: \textit{how can we design an efficient selection mechanism to evolve the population toward greater diversity and higher performance?}

\vspace{-0.5em}
\subsection{Niching-based Selection}\label{sec:niching}
\vspace{-0.5em}
The aforementioned challenge is one that prior agentic automation methods have struggled with: their single-objective optimization approaches often result in increasingly complex workflows, lengthier evaluation/test, and significantly higher API costs~\citep{zhang2024aflow}. For instance, running ADAS~\citep{hu2024adas} on ARC benchmark~\citep{chollet2019arc} with \llmname{gpt-3.5-turbo-0125} incurs cost up to \(\$300\) USD. In contrast, \ourmethod introduces an efficient \textit{niching-based workflow selection} mechanism, guiding evolution across multiple domains and complexity.

\textit{Niching}, in our context, analogous to previous niching EAs~\citep{white2023nas1000}, refers to clusters of similar individuals where environmental selection is conducted. To determine the \textit{niching area} for a new individual \({\mathcal{G}_\circledcirc^{(t)}}\), we compute it based on cost and utility tags as follows:
\begin{equation}
\small
\label{eq:define-niching}
\begin{gathered}
\mathbf{P}^{NA}=\{\mathcal{G}_{q1}, \cdots, \mathcal{G}_{qE}\} = \operatorname{TopK}\left(\{-\operatorname{Rank}(\mathcal{G}_i)\}_{i=1}^N, E\right),\\
\begin{aligned}
  \operatorname{Rank}(\mathcal{G}_i) 
  &= \operatorname{Rank}_{\mathcal{S}}(\mathcal{G}_i) + \operatorname{Rank}_{c}(\mathcal{G}_i)\\
  &= \text{Index}\left(\mathcal{G}_i, \text{Sort}(\{\mathcal{S}({\mathcal{G}_\circledcirc^{(t)}}, \mathcal{G}_j)\}_{j=1}^N)\right) \\
    &+\text{Index}\left(\mathcal{G}_i, \text{Sort}(\{|c({\mathcal{G}_\circledcirc^{(t)}}) - c(\mathcal{G}_j)\}_{j=1}^N)\right) \\
    \end{aligned}
\end{gathered}
\end{equation}
where \(\mathbf{P}^{NA}\) denotes the identified niching area comprising \(E\) individuals. The function \(\operatorname{Rank}(\cdot)\) computes the approximate ranking of an individual \(\mathcal{G}_i\) relative to the new individual, which is determined by the cost similarity rank \(\operatorname{Rank}_c(\cdot)\) and the tag-based similarity rank \(\operatorname{Rank}_\mathcal{S}(\cdot)\). Subsequently, the parents, offspring, and workflows in the niching area are executed for query \(q_t\), and their records are updated as follows:
\begin{equation}
\begin{aligned}
c^{(t)}(\mathcal{G}_i) &= {1}/{t_i'}\left(c^{(t-1)}(\mathcal{G}_i) \cdot t_i' + c(\mathcal{G}_i \mid q_t)\right), \\
p^{(t)}(\mathcal{G}_i) &= {1}/{t_i'}\left(p^{(t-1)}(\mathcal{G}_i) \cdot t_i' + p(\mathcal{G}_i \mid q_t)\right),\\
\mathcal{G}_i &\in \mathbf{P}^{NA}\cup \{\mathcal{G}_{ti}\}_{i=1}^K \cup \{{\mathcal{G}_\circledcirc^{(t)}}\},
\end{aligned}
\end{equation}
where \(c(\mathcal{G} \mid q)\) measures the economical cost incurred by workflow \(\mathcal{G}\) in addressing query \(q\), \(t_i'\) records the number of times workflow \(\mathcal{G}_i\) being executed up to iteration \(t\), \(c^{(t)}(\mathcal{G}_i)\) tracks the cumulative cost of \(\mathcal{G}_i\). \(p(\cdot)\) follows a similar formulation for performance metrics. Finally, the environmental selection is performed within the niching area by calculating the fitness value as follows:  
\begin{equation}\label{eq:fitness}
\mathcal{F}(\mathcal{G}) = \sum_{\mathcal{G} \in \mathbf{P}^{NA} \cup \{\mathcal{G}_\circledcirc^{(t)}\}} \left(\exp \frac{\mathbf{I}(\mathcal{G}, \mathcal{G}_\circledcirc^{(t)})}{\varphi \cdot \mathbf{I}^{\max}}\right),
\end{equation}  
where \(\mathbf{I}(\cdot, \cdot)\) is a Pareto dominance-preserving binary indicator. If individual \(\mathcal{G}_1\) dominates \(\mathcal{G}_2\), \textit{i.e.}, \(c(\mathcal{G}_1) < c(\mathcal{G}_2)\) and \(p(\mathcal{G}_1) < p(\mathcal{G}_2)\) (aligning with our objective in \Cref{eq:obj_1}), then \(\mathbf{I}(\mathcal{G}_1, \mathcal{G}_2) < \mathbf{I}(\mathcal{G}_2, \mathcal{G}_1)\). The maximum absolute indicator value, \(\mathbf{I}^{\max}\), is defined as $
\mathbf{I}^{\max} = \max_{\mathcal{G}_1, \mathcal{G}_2 \in \mathbf{P}^{NA} \cup \{\mathcal{G}_\circledcirc^{(t)}\}} |\mathbf{I}(\mathcal{G}_1, \mathcal{G}_2)|$.  
The scaling factor \(\varphi\) is set to \(0.05\), following established practices~\citep{zitzler2004indicator}. A smaller fitness value in \Cref{eq:fitness} corresponds to a better individual. The worst-performing workflow, \(\mathcal{G}^{\text{worst}}\), which has the largest fitness value in \(\mathbf{P}^{NA} \cup \{\mathcal{G}_\circledcirc^{(t)}\}\), will be eliminated from the population.

\vspace{-0.5em}
\subsection{Discussion}
\vspace{-0.6em}
The evolution process of \ourmethod operates on a query-by-query basis, continuously evolving, mutating, and niching-selecting workflows in response to incoming queries. This iterative process gradually produces a Pareto set of agentic workflows with varying complexity and superior performance. The overall algorithmic procedure is summarized in \Cref{app:alg}, with notations clarified in \Cref{app:notation}.

\vspace{-0.5em}

\begin{table*}[!t]
\centering
\caption{Performance comparison with single agent, hand-craft multi-agent systems, and automated agentic workflows. The base LLM is consistently set as \llmname{get-4o-mini} for all baselines. We \textbf{bold} the best results and \underline{underline} the runner-ups.}
\label{tab:rq1_homo}
\renewcommand\tabcolsep{9pt}
\renewcommand\arraystretch{1.1}
  
\resizebox{\linewidth}{!}{
\begin{tabular}{l|ccccccc}
\Xhline{1.2pt}
\rowcolor{CadetBlue!20} 
{\textbf{Method}} & \textbf{GSM8K} & \textbf{MATH} & \textbf{MultiArith} & \textbf{HumanEval} & \textbf{MBPP}  & \textbf{ALFWorld}  & {\textbf{Avg.}} \\
\Xhline{1.2pt}
Vanilla  & $87.45$ & $46.29$ & $96.85$ & $87.08$ & $71.83$ & $38.71$ & $71.37$ \\
\hline

\rowcolor{gray!10}CoT~\citep{cot} & $87.10$\blue{0.35} & $46.40$\red{0.11} & $96.31$\blue{0.54} & $88.13$\red{1.05} & $71.83$\blue{0.00} & $39.92$\red{1.21} & $71.62$\red{0.25} \\

ComplexCoT~\cite{fu2022complexity}  & $86.89$\blue{0.56} & $46.53$\red{0.24} & $96.70$\blue{0.15} & $87.49$\red{0.41} & $72.36$\red{0.53} & $41.68$\red{2.97} & $71.94$\red{0.57} \\

\rowcolor{gray!10}SC (CoT$\times 5$)~\citep{wang2023selfconsistency}  & $87.57$\red{0.12} & $47.91$\red{1.62} & $96.58$\blue{0.27} & $88.60$\red{1.52} & $73.60$\red{1.77}  & $40.55$\red{1.84} & $72.47$\red{1.10}  \\

\hline

MultiPersona~\citep{multi-persona} & $87.50$\red{0.05} & $45.43$\blue{0.86} & $97.49$\red{0.64} & $88.32$\red{1.24} & $73.19$\red{1.36} & $39.10$\red{0.39} & $71.84$\red{0.47} \\

\rowcolor{gray!10}LLM-Debate~\citep{arXiv2023_MultiAgent-Debate} & $89.47$\red{2.02} & $48.54$\red{2.25} & $97.33$\red{0.48} & $88.68$\red{1.60} & $70.29$\blue{1.54}  & $44.68$\red{5.97} & $73.17$\red{1.80} \\

LLM-Blender~\citep{blender} & $88.35$\red{0.90} & $46.92$\red{0.63} & $97.29$\red{0.44} & $88.80$\red{1.72} & $77.05$\red{5.22}&  $43.79$\red{5.08} & $73.70$\red{2.33} \\

 \rowcolor{gray!10}DyLAN~\citep{arXiv2023_Dynamic-LLM-Agent} & $89.98$\red{2.53} & $48.63$\red{2.34} & $97.12$\red{0.27} & $90.42$\red{3.34} & $77.30$\red{5.47} & $53.32$\red{14.61} & $76.13$\red{4.76} \\ 

AgentVerse~\citep{chen2023agentverse} & $89.91$\red{2.46} & $47.35$\red{1.06} & $97.50$\red{0.65} & $89.29$\red{2.21} & $74.28$\red{2.45} & $45.03$\red{6.32} & $73.89$\red{2.52} \\ 

\rowcolor{gray!10}MacNet~\citep{qian2024scaling} & $87.95$\red{0.50} & $45.18$\blue{1.11} & $96.03$\blue{0.82} & $84.57$\blue{2.51} & $65.28$\blue{6.55} &  $43.66$\red{4.95} & $70.45$\blue{0.92} \\ 

\hline

AutoAgents~\citep{chen2023autoagents} & $87.69$\red{0.24} & $45.32$\blue{0.97} & $96.42$\blue{0.43} & $87.64$\red{0.56} & $71.95$\red{0.12} & $46.15$\red{7.44} & $72.53$\red{1.16} \\

\rowcolor{gray!10}GPTSwarm~\citep{zhuge2024gptswarm}  & $89.14$\red{1.69} & $47.88$\red{1.59} & $96.79$\blue{0.06} & $89.32$\red{2.24} & $77.43$\red{5.60} &  $53.19$\red{14.48} & $75.63$\red{4.26}  \\

ADAS~\citep{hu2024adas}  & $86.12$\blue{1.33} & $43.18$\blue{3.11} & $96.02$\blue{0.83} & $84.19$\blue{2.89} & $68.13$\blue{3.70} & $47.66$\red{8.95} & $70.88$\blue{0.49}  \\

\rowcolor{gray!10}AgentSquare~\citep{shang2024agentsquare} & $87.62$\red{0.17} & $48.51$\red{2.22} & \underline{$97.77$}\red{0.92} & {$89.08$}\red{3.00} & $78.46$\red{6.63} & \underline{$66.42$}\red{27.71} & $78.14$\red{6.77} \\

AFlow~\citep{zhang2024aflow}  & \underline{$91.16$}\red{3.71} & \underline{$51.28$}\red{3.31} & $96.22$\blue{0.63} & \underline{$90.93$}\red{3.85} & \underline{$81.67$}\red{9.84} & $59.16$\red{20.45} &    $78.40$\red{7.03}  \\

\hline

\rowcolor{gray!10}\textbf{\ourmethod (Ours)}  & {$\mathbf{92.90}$}\red{4.85} & \textbf{$\mathbf{57.70}$}\red{11.41} & \textbf{$\mathbf{98.80}$}\red{1.95} & \textbf{$\mathbf{92.85}$}\red{5.77} & \textbf{$\mathbf{84.50}$}\red{10.34} & \textbf{$\mathbf{68.57}$}\red{29.86} &\textbf{$\mathbf{82.55}$}\red{11.18} \\

\Xhline{1.2pt}
\end{tabular}
}
\vspace{-1.5em}
\end{table*}

\begin{table*}[!t]
\centering
\caption{Heterogeneous experiments on MATH and MBPP. ``DyLAN\textsubscript{Qwen}'' indicates that only \llmname{Qwen-2.5-72b} was used to optimize DyLAN. For comparison, we included results from \llmname{o1-preview}, although \ourmethod exclusively utilized four open-source LLMs. We shade the values of the lowest overall cost, the lowest inference token, and the highest performance for both single agents and workflows.}
% \vspace{-0.1em}
\setlength\tabcolsep{6pt}
\resizebox{1\textwidth}{!}{
\begin{tabular}{llcccccccccc}
\Xhline{1.2pt} % & \multirow{2}{*}{Methods}
\multicolumn{2}{c}{\multirow{2}{*}{Model}} & \multicolumn{5}{c}{\textbf{MATH}
} & \multicolumn{5}{c}{\textbf{MBPP}}    \\
\cmidrule(lr){3-7} \cmidrule(lr){8-12}
 & & \makecell{Training\\cost ($10^{-3} \$$)}& {\makecell{Inference\\cost ($10^{-3} \$$)}} & \makecell{Overall\\cost ($10^{-3} \$$)}& \makecell{Inference\\token}& {\makecell{Acc.\\(\%)}} & \makecell{Training\\cost ($10^{-3} \$$)}& {\makecell{Inference\\cost ($10^{-3} \$$)}} & \makecell{Overall\\cost ($10^{-3} \$$)}& \makecell{Overall\\token}& {\makecell{pass@1\\(\%)}} \\
% \Xhline{1.pt}
\hline
\parbox[t]{2mm}{\multirow{5}{*}{\rotatebox[origin=c]{90}{Single}}} & Llama-3.1-70b & - & $24.50$ & $24.50$ & $90,678$ & $31.93\%$ & - & $10.67$ & $10.67$ & $38,653$ & $65.11\%$  \\
 & Qwen-2.5-72b & - & $32.30$ & $32.30$ & $85,436$ & $63.80\%$ & - & $9.18$ & \cellcolor{gray!25}$9.18$ & \cellcolor{gray!25}$24,253$ & $69.76\%$\\
 & Deepseek-V2.5 & - & $25.89$  & $25.89$  & $98,986$ & $41.17\%$ & - & $11.93$ & $11.93$ & $44,589$ & $76.74\%$    \\
  & Hermes-3-70b  & - & $18.11$ & \cellcolor{gray!25}$18.11$ & \cellcolor{gray!25} $68,994$ & $22.60\%$ & - & $9.49$ & $9.49$ & $30,328$ & $63.28\%$  \\
  \cmidrule(lr){2-12}
   & o1-preview  & - & $7840.51$ & $7840.51$ & $186,701$ &\cellcolor{gray!25} $70.20\%$ & - &$3209.44$ & $3209.44$ & $81,334$ & \cellcolor{gray!25}$89.65\%$ \\
  \hline
  \parbox[t]{2mm}{\multirow{8}{*}{\rotatebox[origin=c]{90}{Homogeneous}}} & AFlow\textsubscript{Llama} & $653.97$ & $1304.07$ & $1958.05$ & $6,054,698$ & $36.97\%$ & $383.40$ & $356.88$ & $740.29$ & $1,510,058$ & $67.42\%$\\
 & AFlow\textsubscript{Qwen} &  $1223.46$ & $2622.63$ & $3846.10$ & $8,614,237$ & $66.38\%$ & $824.48$ & $773.63$ & $1598.11$ & $2,258,279$ & $80.84\%$ \\
 & AFlow\textsubscript{Deepseek} & $815.96$ & $1945.90$ & $2761.86$ & $8,693,402$ &  $48.65\%$ & $456.96$ & $418.33$ & $875.29$ & $1,733,829$ & $79.14\%$  \\
  & {AFlow\textsubscript{Hermes}} & $572.09$ & $1045.20$ & $1617.29$ & $4,886,371$ & $32.14\%$ & $353.04$ & $339.71$ & $692.75$ & $1,289,901$ & $66.13\%$ \\
  \cmidrule(lr){2-12}
  & DyLAN\textsubscript{Llama} & $9317.07$ & $2676.03$ & $11993.10$ & $11,258,530$ & $38.19\%$ & $5817.09$ & $966.12$ & $6783.21$ & $4,879,526$ & $69.92\%$ \\
   & DyLAN\textsubscript{Qwen} & $12847.72$ & $4015.97$ & $16863.69$ & $15,242,982$ & $64.17\%$ & $7491.78$ & $1480.94$ & $8972.72$ & $3,386,087$ & $75.63\%$\\
 & DyLAN\textsubscript{Deepseek}  & $10388.54$ & $2375.11$ &  $12763.64$ & $13,282,450$ & $46.20\%$ & $6209.41$ & $1084.34$ & $7293.45$ & $4,296,199$ & $80.13\%$\\
  & DyLAN\textsubscript{Hermes} & $7103.88$ & $2106.35$ & $9210.23$ & $8,129,786$  & $30.14\%$ & $3965.38$ & $714.55$ & $4679.93$ & $4,150,887$ & $65.29\%$\\
  \hline
  & \ourmethod & $459.24$ & $513.34$ &\cellcolor{gray!25} $972.58$ & \cellcolor{gray!25}$1,660,284$ & \cellcolor{gray!25}$72.90\%$ & $479.10$ & $286.05$ & \cellcolor{gray!25}$565.15$ & \cellcolor{gray!25} $8,193,669$ & \cellcolor{gray!25}$87.62\%$ \\
 
 % & \ourmethod & $40.00\%$ & \cellcolor{gray!25}$1397.59$ & \cellcolor{gray!25}$3.36 (1.27\times)$& $43.00\%$ & \cellcolor{gray!25}$3890.80$ & \cellcolor{gray!25}$4.48 (1.35\times)$& $35.00\%$ &\cellcolor{gray!25} $3092.81$ & \cellcolor{gray!25}$4.82 (1.29\times)$\\

% \midrule
\Xhline{1.2pt}
\end{tabular}}\label{tab:heterogeneous}
\vspace{-1.3em}
\end{table*}

\section{Experiments}
\vspace{-0.5em}
\subsection{Experiment Setup}\label{sec:exp-setup}
\vspace{-0.4em}
\paragraph{Tasks and Benchmarks.} We evaluate \ourmethod on six public benchmarks covering four domains: \textbf{(1) math reasoning}, GSM8K~\citep{gsm8k}, MATH~\citep{hendrycksmath2021}, and MultiArith~\cite{roy2016solving};  \textbf{(2) code generation}, HumanEval~\citep{human-eval} and MBPP~\citep{austin2021mbpp}); \textbf{(3) embodied}, ALFWorld~\citep{shridhar2021alfworld}. For the MATH benchmark, we follow \cite{hong2024datainterpreter} in selecting a harder subset (617 problems). The dataset statistics and splits are in \Cref{app:dataset}.
\vspace{-0.8em}
\paragraph{Baselines.}  We compare \ourmethod with two series of agentic baselines: \textbf{(1) manually designed workflows}, including Chain-of-Thought~\cite{cot}, ComplexCoT~\cite{fu2022complexity}), Self-Consistency (SC)~\cite{wang2023selfconsistency},  LLM-Debate~\citep{arXiv2023_MultiAgent-Debate}, LLM-Blender~\cite{blender}, DyLAN~\citep{arXiv2023_Dynamic-LLM-Agent}, AgentVerse~\citep{chen2023agentverse} and MacNet~\citep{qian2024scaling}; \textbf{(2) autonomous workflows}, including GPTSwarm~\citep{zhuge2024gptswarm}, AutoAgents~\citep{chen2023autoagents}, ADAS~\citep{hu2024adas}, AgentSquare~\citep{shang2024agentsquare} and AFlow~\citep{zhang2024aflow}. Detailed baseline setups are in \Cref{app:baselines}.

\vspace{-0.8em}
\paragraph{LLM Backbones.} We leverage one closed-source model, \llmname{gpt-4o-mini-0718}, along with four open-source models: \llmname{llama-3.1-70b}, \llmname{Qwen-2-72b}, \llmname{Deepseek-V2.5}, and \llmname{Hermes-3-70b}. LLMs are accessed via APIs, with the temperature set to $1$.

\vspace{-0.8em}
\paragraph{Parameter Configuration.} We select the following operators to initialize the feasible space of operator nodes: CoT, LLM-Debate, Take-a-step-back, Self-consistency, Self-Refine, Ensemble, ReAct, and ExpertPrompting. Detailed instructions are in \Cref{app:operator}. The function $\mathbf{v}(\cdot)$ adopts all-MiniLM-L6-v2~\citep{wang2020minilm}. The number of parent workflows in \Cref{eq:select-parent} is set as  $K=3$, and the number of utility indicator tags in \Cref{eq:tag} is set as $\kappa=5$. The population size $N$ is $15$, and $E=5$ in \Cref{eq:define-niching}.

\begin{figure*}[t]
\centering
\includegraphics[width=1.0\linewidth]{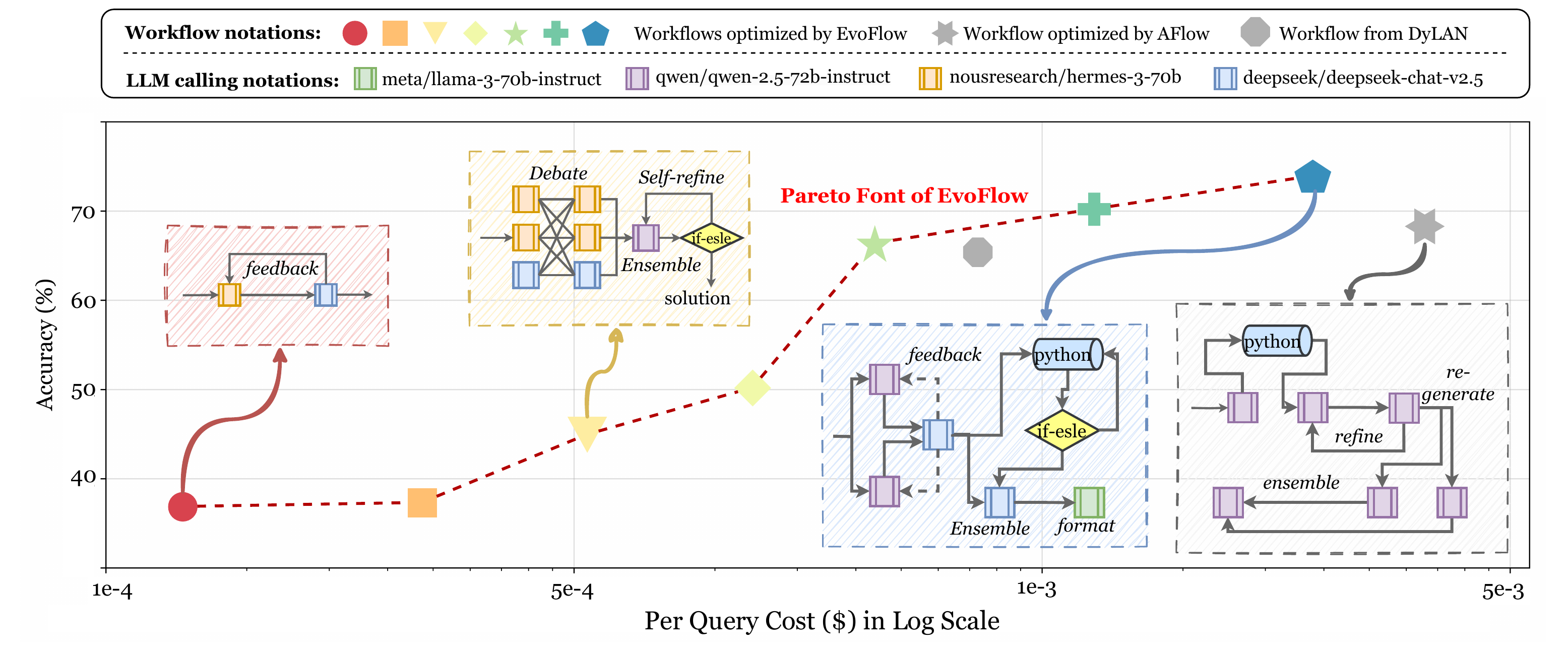}
\vspace{-1cm}
\caption{The cost-performance plane of workflows from \ourmethod, DyLAN, and AFlow.
% The red and blue lines represent the ROC-AUC  (left $y$-axis) and the spectral preservation ratio\footnotemark of the sparse subgraph (right $y$-axis), respectively. The horizontal red dashed line indicates the original performance. 
%It is observed that with increasing sparsity, both sparsifiers result in decreased performance and spectral preservation, yet the decline with Local Degree is more gradual.
}
\vspace{-1.2em}
\label{fig:pareto}
\end{figure*}

\begin{figure}[t]
\centering
\includegraphics[width=1.0\linewidth]{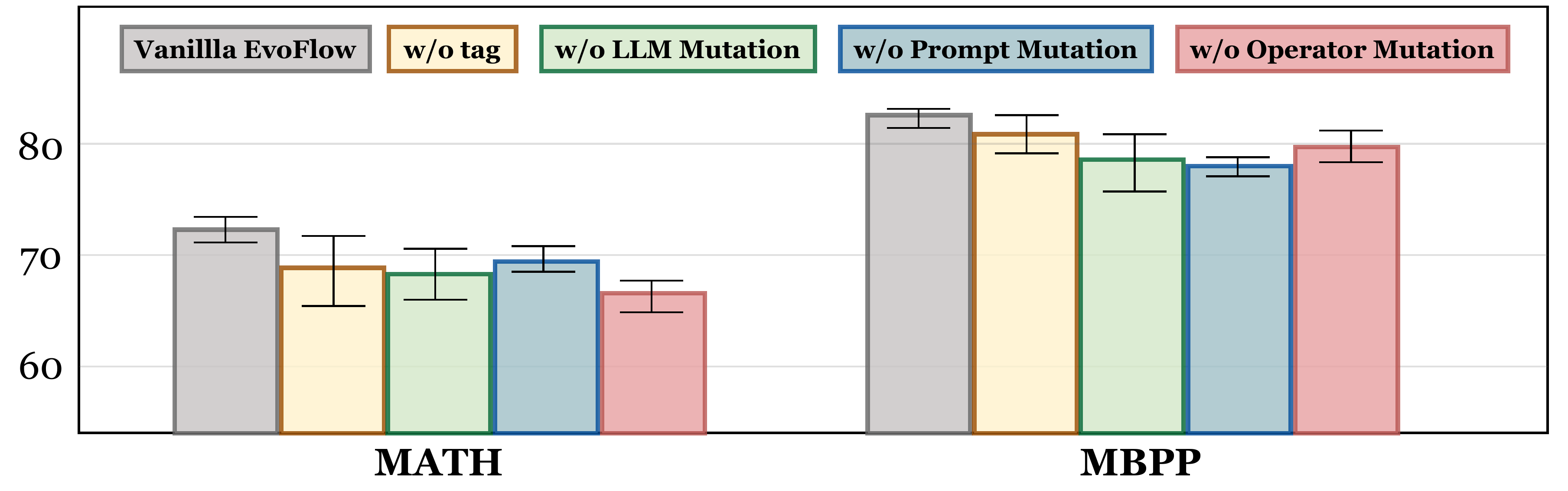}
\vspace{-0.8cm}
\caption{The ablation study of \ourmethod.
% The red and blue lines represent the ROC-AUC  (left $y$-axis) and the spectral preservation ratio\footnotemark of the sparse subgraph (right $y$-axis), respectively. The horizontal red dashed line indicates the original performance. 
%It is observed that with increasing sparsity, both sparsifiers result in decreased performance and spectral preservation, yet the decline with Local Degree is more gradual.
}
\vspace{-1.2em}
\label{fig:ablation}
\end{figure}

\begin{figure}[t]
\centering
\includegraphics[width=1.0\linewidth]{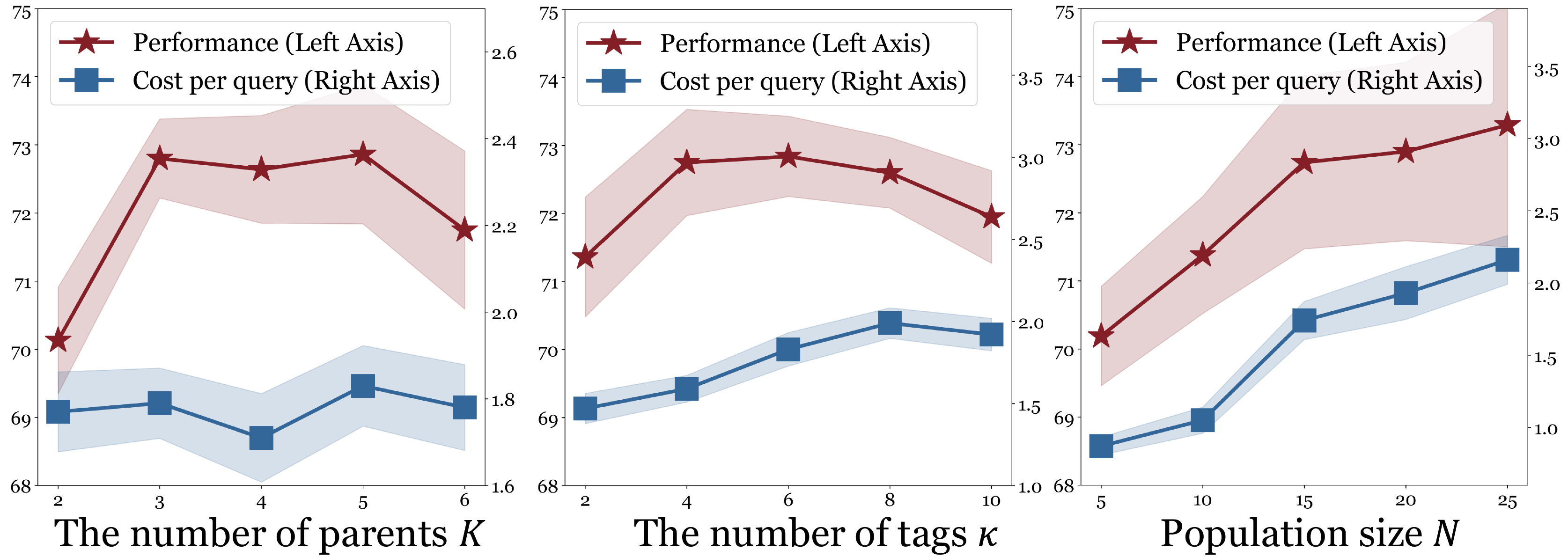}
\vspace{-0.8cm}
\caption{The parameter sensitivity analysis of \ourmethod. The unit of cost per query (right) and performance (left) is $10^{-3}\cdot\$$ and accuracy (\%), respectively.
}
\vspace{-1.5em}
\label{fig:sensi}
\end{figure}

\vspace{-0.6em}
\subsection{Performance Analysis}
\vspace{-0.7em}
We present via three experimental settings: \textbf{\ding{110} homogeneous setting}, where all methods, including \ourmethod, are equipped with a unified LLM backbone; \textbf{\ding{110} heterogeneous setting}, where \ourmethod is assigned an LLM pool and optimizes a heterogeneous agentic workflow population; \textbf{\ding{110} cross-domain setting}, where we mix up multiple cross-domain datasets for training. The analysis is as follows:

\vspace{-1em}
\paragraph{Homogeneous Performance} \Cref{tab:rq1_homo} demonstrates that \ourmethod outperforms existing hand-crafted or automated agentic workflows across six benchmarks. Specifically, on the MATH benchmark, it exceeds vanilla \llmname{gpt-4o-mini} by $11.41\%$ and surpasses the SOTA baseline AFlow by $6.42\%$. On the embodied benchmark ALFWorld, \ourmethod achieves the optimal $68.57\%$, outperforming the second-best AgentSquare by $2.15\%$.

\vspace{-0.9em}
\paragraph{Heterogeneous Performance} \llmname{qwen-2.5-72b} exhibits the best performance among the four open-source models, but even with the sophisticated optimization from AFlow, it only shows a $2.58\%$ improvement on MATH, still trailing behind the powerful \llmname{o1-preview} by $3.82\%\downarrow$; however, \ourmethod, through the collective assembly and evolution of the four open-source models, surpasses \llmname{o1-preview} by $2.7\%$. More importantly, the overall cost of \ourmethod is merely $12.4\%$ of that of \llmname{o1-preview}. This clearly illustrates both the necessity and potential of optimizing LLM-heterogeneous workflows.

\vspace{-1em}
\paragraph{Cross-domain Performance} We also include a cross-domain optimization setting, where training sets from different domain datasets are concatenated to assess whether an agentic automation method can optimize reasonable workflows across domains. 
%The detailed explanations are in \Cref{app:exp-result}.
As shown in \Cref{tab:dylan_comparison}, cross-domain optimization challenges many existing baselines: vanilla \llmname{Deepseek-V2.5} achieves $41.17\%$ on MATH, and GPTSwarm improves it by $4.19\%$ when optimized solely for MATH. However, joint optimization on MATH+MBPP results in a negative gain, reducing performance to $39.18\%$. Other methods like DyLAN and AFlow also suffer from the same issue. In contrast, \ourmethod successfully benefits from cross-domain training on MBPP, improving from $87.62\%$ to $88.35\%$, a result attributed to the optimization of the workflow population rather than a single individual.

\vspace{-0.8em}
\subsection{Cost Analysis \& Case Study}
\vspace{-0.5em}
We demonstrate the resource-friendly nature of \ourmethod's agentic automation system across three dimensions: \textbf{training/inference API costs} and \textbf{token consumption}. As shown in \Cref{tab:heterogeneous}, optimizing AFlow\textsubscript{Qwen} on MATH incurs a training cost of $1.22\$$ and an inference cost of $2.62\$$. In comparison, \ourmethod requires only $37.5\%$ of the training cost and $19.5\%$ of the inference cost: this is because \ourmethod’s workflows do not continuously rely on the most expensive \llmname{qwen-2.5-72b}, and instead opt for more economical models such as \llmname{llama-3.1} or \llmname{herme-3-70b} when appropriate, without compromising performance. %This highlights the superiority of heterogeneous searching.

We further visualize the optimized heterogeneous population of \ourmethod in \Cref{fig:pareto}. It can be observed that \ourmethod forms a Pareto front in this performance-cost plane. The population begins with simple and inexpensive workflows, consisting solely of basic I/O and self-refine (accuracy@$38.7\%$, cost@$0.00018\$$), and progresses to more complex workflows incorporating multi-agent debate. The most high-performing workflows include iterative generation and ensembling, but at the cost of higher per-query token consumption (accuracy@$72.57\%$, cost@$0.0037\$$). This highlights \ourmethod's query-aware paradigm: for simple queries, it intelligently selects economical workflows for rapid completion, while for more complex ones, it leverages sophisticated workflows to address the increased demands.

\vspace{-0.4em}
\subsection{Framework Analysis}
\vspace{-0.4em}
\paragraph{Ablation Study}  We perform an ablation study on four variants of \ourmethod: \textbf{\textit{w/o} tag}, where tag-based retrieval is removed and replaced with random selection in \Cref{eq:select-parent}; \textbf{\textit{w/o} LLM mutation}, where \Cref{eq:llm-mutation} is discarded; \textbf{\textit{w/o} prompt mutation}, where \Cref{eq:prompt-mutation} is removed; and \textbf{\textit{w/o} operator mutation}, where \Cref{eq:operator-mutation} is discarded. The results from \Cref{fig:ablation} reveal that removing tag-based retrieval and LLM mutation consistently leads to performance degradation and greater variance. This is because \ourmethod becomes heavily influenced by random parent workflow selection and the LLM backbone chosen during individual initialization. Removing operator mutation, which eliminates the potential for creating new, creative operators, results in a performance drop of $3.5\%\sim7.3\%$.

\vspace{-1em}
\paragraph{Sensitivity Analysis}
We perform an ablation study on three key parameters of \ourmethod: the number of selected parents $K$ in \Cref{eq:select-parent}, the number of tags per individual $\kappa$ in \Cref{eq:tag}, and the population size $N$. As shown in \Cref{fig:sensi}, (1) both too small and large $K$ result in performance degradation, likely because a small $K$ reduces offspring diversity, while a large $K$ challenges the LLM's ability to aggregate multiple workflows; (2) increasing the population size $N$ consistently improves performance, with a gain of $3.1\%$ from $N=5$ to $N=25$. However, a larger population also increases complexity, with the per-query cost rising from $8e-4$ to $2e-3$. Balancing cost-effectiveness, we set $N=15$ across all experiments.

\vspace{-0.5em}
\section{Conclusion}
\vspace{-0.7em}
In this paper, we shift the paradigm of autonomous multi-agent workflow search from single-objective to cost-effectiveness-driven multi-objective optimization. Building on niching-based evolutionary algorithms, we propose \ourmethod, an autonomous framework that evolves a population of heterogeneous, complexity-adapted agentic workflows. Extensive experiments across six benchmarks demonstrate the superior performance of \ourmethod with significantly lower token costs.

% \subsection{Theorems and such}
% The preferred way is to number definitions, propositions, lemmas, etc. consecutively, within sections, as shown below.
% \begin{definition}
% \label{def:inj}
% A function $f:X \to Y$ is injective if for any $x,y\in X$ different, $f(x)\ne f(y)$.
% \end{definition}
% Using \cref{def:inj} we immediate get the following result:
% \begin{proposition}
% If $f$ is injective mapping a set $X$ to another set $Y$, 
% the cardinality of $Y$ is at least as large as that of $X$
% \end{proposition}
% \begin{proof} 
% Left as an exercise to the reader. 
% \end{proof}
% \cref{lem:usefullemma} stated next will prove to be useful.
% \begin{lemma}
% \label{lem:usefullemma}
% For any $f:X \to Y$ and $g:Y\to Z$ injective functions, $f \circ g$ is injective.
% \end{lemma}
% \begin{theorem}
% \label{thm:bigtheorem}
% If $f:X\to Y$ is bijective, the cardinality of $X$ and $Y$ are the same.
% \end{theorem}
% An easy corollary of \cref{thm:bigtheorem} is the following:
% \begin{corollary}
% If $f:X\to Y$ is bijective, 
% the cardinality of $X$ is at least as large as that of $Y$.
% \end{corollary}
% \begin{assumption}
% The set $X$ is finite.
% \label{ass:xfinite}
% \end{assumption}
% \begin{remark}
% According to some, it is only the finite case (cf. \cref{ass:xfinite}) that is interesting.
% \end{remark}
%restatable

\section*{Impact Statement}

\paragraph{Ethical impacts.} We affirm that our proposed \ourmethod method poses no ethical concerns in terms of its motivation, design, experiments, and data usage. The method is built with a focus on fostering advancements in multi-agent systems, ensuring its responsible contribution to scientific research and the development of more efficient and customized solutions.
\vspace{-0.5em}
\paragraph{Expected societal implications.} \ourmethod presents a transformative approach to multi-agent systems by promoting the automation of heterogeneous agent workflows, optimizing for both performance and cost. By facilitating the use of diverse models for task-specific solutions, it offers new avenues for deploying more adaptive and scalable systems in various practical domains.  
% In the unusual situation where you want a paper to appear in the
% references without citing it in the main text, use \nocite
% \nocite{langley00}

\bibliography{example_paper}
\bibliographystyle{icml2025}

%%%%%%%%%%%%%%%%%%%%%%%%%%%%%%%%%%%%%%%%%%%%%%%%%%%%%%%%%%%%%%%%%%%%%%%%%%%%%%%
%%%%%%%%%%%%%%%%%%%%%%%%%%%%%%%%%%%%%%%%%%%%%%%%%%%%%%%%%%%%%%%%%%%%%%%%%%%%%%%
% APPENDIX
%%%%%%%%%%%%%%%%%%%%%%%%%%%%%%%%%%%%%%%%%%%%%%%%%%%%%%%%%%%%%%%%%%%%%%%%%%%%%%%
%%%%%%%%%%%%%%%%%%%%%%%%%%%%%%%%%%%%%%%%%%%%%%%%%%%%%%%%%%%%%%%%%%%%%%%%%%%%%%%
\newpage
\appendix
\onecolumn

\section{Notations}
\label{app:notation}

We present a comprehensive review of the commonly used notations and their definitions in \Cref{tab:notations}.

\begin{table}[ht]
\centering
\caption{Notation and Definitions}
\begin{tabular}{ll}
\toprule
\textbf{Notation} & \textbf{Definition} \\
\midrule
$I_i$ & An LLM-invoking node \\
$P_i$ & The prompt content of $I_i$ \\
$\mathcal{P}$ & The feasible prompt space \\
$M_i$ & The base LLM invoked by $I_i$ \\
$\tau_i$ & Temperature of $M_i$ \\
$\mathcal{M} = \{M_1, \cdots, M_{|\mathcal{M}|}\}$ & LLM pool \\
$\mathcal{I} = \mathcal{M} \times \mathcal{P} \times \mathbb{R}_{[0,1]}$ & The feasible space for invoking nodes \\
$O_j = (\mathcal{I}^o_j, \mathcal{E}^o_j)$ & An operator node composed of multiple invoking nodes \\
$\mathcal{I}^o_j = \{I_1, \dots, I_n\}$ & The selected invoking nodes in $O_j$ \\
$\mathcal{E}^o_j \subseteq \mathcal{I}^o_j \times \mathcal{I}^o_j$ & The connectivity of operator nodes in $O_j$ \\
$\mathcal{O}$ & The feasible space of operator nodes \\
$\mathcal{G} = (\mathcal{O}^S, \mathcal{E}^a) = (\mathcal{I}^S, \mathcal{E}^o)$ & An agentic workflow \\
$\mathcal{O}^S \in \mathcal{O}$ & A subset of operator nodes used in $\mathcal{G}$ \\
$\mathcal{I}^S \in \mathcal{I}$ & A subset of invoking nodes selected in $\mathcal{G}$ \\
$u(\mathcal{G}, T)$ & An evaluator function assessing $\mathcal{G}$'s performance in task domain $T$ \\
$c(\mathcal{G}, T)$ & An evaluator function assessing $\mathcal{G}$'s cost in task domain $T$ \\
$\mathcal{G}^*$ & The best workflow searched by baseline methods \\
$\mathcal{G}^\star$ & The Pareto-optimal set of agentic workflows balancing cost and performance \\
$\mathbf{P}^{(t)} = \{\mathcal{G}_1, \mathcal{G}_2, \cdots, \mathcal{G}_N\}$ & A population of $N$ agentic workflows at the $t$-th iteration \\
$\varkappa^k_i$ & The $i$-th tag of workflow $\mathcal{G}_k$ \\
$\mathbf{v}(\cdot)$ & Text embedding function \\
$\mathcal{S}(\mathcal{G} | q)$ & The similarity score of workflow $\mathcal{G}$ with respect to query $q$ \\
$\mathcal{G}_\circ^{(t)}$ & The generated offspring workflow at the $t$-th iteration \\
$\mu^l(\cdot)$ & LLM mutation function \\
$\mu^p(\cdot)$ & Prompt mutation function \\
$\mu^o(\cdot)$ & Operator mutation function \\
$\tilde{\mathcal{G}}$ & A mutated workflow \\
$\mathcal{G}^{(t)}_\circledcirc$ & A mutated workflow at the $t$-th iteration \\
$\mathbf{P}^{NA} = \{\mathcal{G}_{q1}, \cdots, \mathcal{G}_{qE}\}$ & The identified niche area comprising $E$ individuals \\
$c^{(t)}(\mathcal{G}_i)$ & The cumulative cost of $\mathcal{G}_i$ at the $t$-th iteration \\
$u^{(t)}(\mathcal{G}_i)$ & The cumulative performance of $\mathcal{G}_i$ at the $t$-th iteration \\
$\mathbf{I}(\cdot, \cdot)$ & Pareto dominance-preserving binary indicator \\
$\mathcal{F}(\mathcal{G})$ & The fitness value of $\mathcal{G}$ \\
\bottomrule
\end{tabular}
\label{tab:notations}
\end{table}

\section{Algorithm Table}
\label{app:alg}

We conclude the overall algorithm procedure of \ourmethod in \Cref{alg:algo}.
\begin{algorithm}[!t]
\caption{Algorithm workflow of \ourmethod}\label{alg:algo}
\Input{A dataset $\mathcal{D}$ containing training set $\mathcal{D}_\text{train}$ and test set $\mathcal{D}_\text{test}$, Operator set $\mathcal{O}$}
                
\Output{The well-optimized, diverse workflow population $\mathbf{P}^{|\mathcal{D}_\text{train}|}$}

% \Output{Optimized operators $\mathbb{O}$ and distribution $\boldsymbol{pi}$}

\tcc{\textcolor{blue}{Initialize workflow population}}
% Initialize workflow population 
\For{\rm{individual} $k \leftarrow 1$ \KwTo $N$}{
$\mathcal{G}_k \leftarrow \{\mathcal{O}_k, \mathcal{E}^a_k\},  \mathcal{E}^a_k \subseteq \mathcal{O}_k \times \mathcal{O}_k$\Comment*[r]{\textcolor{blue}{Eq.~\ref{eq:init-population}}}
\tcc{\textcolor{blue}{Assigning utility tags}}
$\{\varkappa^k_1, \cdots, \varkappa^k_\kappa\} \leftarrow f_\text{tag}(\mathcal{G}_k)$\Comment*[r]{\textcolor{blue}{Eq.~\ref{eq:tag}}}
}
% $\mathbf{P}^{(0)}=\{\mathcal{G}_1, \mathcal{G}_2, \cdots, \mathcal{G}_N\}$

Obtain initialized population $\mathbf{P}^{(0)}=\{\mathcal{G}_1, \mathcal{G}_2, \cdots, \mathcal{G}_N\}$
% \tcc{ for }

\For{\rm{query} $q_t$ \rm{in} $\mathcal{D}_\text{train}$}{
\tcc{\textcolor{blue}{Retrieve relevant workflows via tag-based similarity}}

Locate parent workflows: $\{\mathcal{G}_{t1}, \cdots, \mathcal{G}_{tK}\} = \operatorname{TopK}\left(\mathcal{S}(\{\mathcal{G}_i\}_{i=1}^N \;|\; q_t), K\right)$\Comment*[r]{\textcolor{blue}{Eq.~\ref{eq:select-parent}}}

\tcc{\textcolor{blue}{Crossover and generate the offspring workflow}}

Evolve the offspring workflow via the Crossover function: $\mathcal{G}_\circ^{(t)}\leftarrow \operatorname{Crossover}(\mathcal{G}_{t1}, \cdots, \mathcal{G}_{tK})$\Comment*[r]{\textcolor{blue}{Eq.~\ref{eq:crossover}}}

Mutate the offspring with three mutation functions: LLM mutation $\mu^l(\cdot)$, Prompt Mutation $\mu^{p}(\cdot)$ and Operator Mutation $\mu^{o}(\cdot)$; obtain the mutated offspring ${\mathcal{G}_\circledcirc^{(t)}}$\Comment*[r]{\textcolor{blue}{\Cref{eq:operator-mutation,eq:prompt-mutation,eq:llm-mutation}}}

\tcc{\textcolor{blue}{Niching-based selection \& elimination}}

Identify the niching area $\mathbf{P}^{NA}$: $\mathbf{P}^{NA}=\{\mathcal{G}_{q1}, \cdots, \mathcal{G}_{qE}\} = \operatorname{TopK}\left(\{-\operatorname{Rank}(\mathcal{G}_i)\}_{i=1}^N, E\right)$\Comment*[r]{\textcolor{blue}{Eq.~\ref{eq:define-niching}}}

\tcc{\textcolor{blue}{Execute parents, offsprings and niching-area workflows}}

\For{\rm{workflow} $\mathcal{G}_i \in \mathbf{P}^{NA}\cup \{\mathcal{G}_{ti}\}_{i=1}^K \cup \{{\mathcal{G}_\circledcirc^{(t)}}$}{

Execute $\mathcal{G}_i$ on query $q$

\tcc{\textcolor{blue}{Record and update its cost}}
$c^{(t)}(\mathcal{G}_i) = {1}/{t_i'}\left(c^{(t-1)}(\mathcal{G}_i) \cdot t_i' + c(\mathcal{G}_i \mid q_t)\right)$

\tcc{\textcolor{blue}{Record and update its performance}}
$p^{(t)}(\mathcal{G}_i) = {1}/{t_i'}\left(p^{(t-1)}(\mathcal{G}_i) \cdot t_i' + p(\mathcal{G}_i \mid q_t)\right)$
}

\For{\rm{workflow} $\mathcal{G}_i \in \mathbf{P}^{NA}\cup \{\mathcal{G}_{ti}\}_{i=1}^K \cup \{{\mathcal{G}_\circledcirc^{(t)}}$}{
\tcc{\textcolor{blue}{Calculate each individual's fitness value}}
Calculate the fitness value $\mathcal{F}(\mathcal{G}) = \sum_{\mathcal{G} \in \mathbf{P}^{NA}} \left(\exp \frac{\mathbf{I}(\mathcal{G}, \mathcal{G}_\circledcirc^{(t)})}{\varphi \cdot \mathbf{I}^{\max}}\right)$

}

Locate the individual with the largest $\mathcal{F}(\mathcal{G})$ as $\mathcal{G}^\text{worst}$

\tcc{\textcolor{blue}{Update the population}}

Update the population $\mathbf{P}^{(t+1)} \leftarrow \mathbf{P}^{(t)} \setminus\mathcal{G}^\text{worst} \cup \mathcal{G}_\circledcirc^{(t)}$

\tcc{\textcolor{blue}{Notably, if the generated workflow performs suboptimally, i.e., it does not Pareto dominate any existing workflows, it is unlikely to be accepted into the population.}}

}
\end{algorithm}
% \vspace{-0.5em}

\section{Optimization Objective}
\label{app:obj}

For a better understanding of the multi-objective optimization in the context of agentic workflows, here, we define some key concepts like dominance, Pareto optimality, and the Pareto set. 

A workflow \(\mathcal{G}_1\) is said to \textbf{dominate} another workflow \(\mathcal{G}_2\) if and only if:  
\begin{equation}
\forall i \in \{1, 2\}, \; f_i(\mathcal{G}_1) \geq f_i(\mathcal{G}_2), \; \text{and} \; \exists i \; \text{such that} \; f_i(\mathcal{G}_1) > f_i(\mathcal{G}_2),
\end{equation}  
where \(f_1(\cdot) = u(\cdot, T)\) represents the utility or performance metric, and \(f_2(\cdot) = -c(\cdot, T)\) denotes the negative system cost. This means \(\mathcal{G}_1\) performs at least as well as \(\mathcal{G}_2\) in all objectives and strictly better in at least one objective.  

A workflow \(\mathcal{G}^\star \in \mathcal{H}(\mathcal{I}, \mathcal{E})\) is considered \textbf{Pareto optimal} if there does not exist any other workflow \(\mathcal{G}' \in \mathcal{H}(\mathcal{I}, \mathcal{E})\) that dominates \(\mathcal{G}^\star\). The collection of all Pareto optimal workflows forms the \textbf{Pareto set}:  
\begin{equation}
\mathcal{G}^*_{\text{PF}} = \left\{\mathcal{G} \in \mathcal{H}(\mathcal{I}, \mathcal{E}) \mid \nexists \mathcal{G}' \in \mathcal{H}(\mathcal{I}, \mathcal{E}), \mathcal{G}' \; \text{dominates} \; \mathcal{G} \right\}.
\end{equation}  

The corresponding objective space of these workflows defines the \textbf{Pareto front (PF)}, which represents the trade-off surface between performance and cost:  
\begin{equation}
\text{PF} = \left\{ \mathbf{f}(\mathcal{G}) = [u(\mathcal{G}, T), -c(\mathcal{G}, T)]^\top \mid \mathcal{G} \in \mathcal{G}^*_{\text{PF}} \right\}.
\end{equation}  

In the context of agentic workflows, identifying Pareto optimal solutions is critical as it enables the selection of workflows that provide the best possible trade-off between task performance and system cost. These solutions ensure that the agents, composed of invoking nodes and operator nodes, operate efficiently while maintaining high utility for the target tasks. Furthermore, the Pareto set provides diverse design options, offering flexibility in adapting workflows to varying operational constraints and objectives.

\section{Operator Repository}\label{app:operator}

In this section, we detail the initialization of operator nodes, which can be categorized into the following seven types:

\begin{enumerate}
    \item \textbf{Chain-of-Thought (CoT).}  
    CoT~\citep{cot} reasoning encourages the LLM to think step by step rather than directly outputting an answer. This approach enhances its capability to solve complex problems through intermediate reasoning steps, improving task handling and providing greater transparency in the decision-making process.

    \item \textbf{LLM-Debate.}  
    LLM-Debate~\citep{arXiv2023_MultiAgent-Debate} allows multiple LLMs to debate, leveraging diverse perspectives to identify better solutions. In practice, we initialize three debaters and permit up to two debate rounds.

    \item \textbf{Take a Step Back.}  
    As proposed by \citet{zheng2023take-a-step-back}, this operator prompts the LLM to first consider the principles underlying the task. By focusing on foundational principles, the model enhances its reasoning and delivers more accurate solutions.

    \item \textbf{Self-Consistency.}  
    Adopting the methodology from \citet{wang2023selfconsistency}, this operator aggregates five CoT reasoning paths and determines the final answer through majority voting.

    \item \textbf{Self-Refine.}  
    Following \citet{NeurIPS2023_Self-Refine}, this operator initially generates an answer using CoT reasoning, then prompts the agent to self-reflect iteratively. We set a maximum of five refinement iterations.

    \item \textbf{Ensemble.}  
    Inspired by LLM-Blender~\citep{blender}, this operator involves three LLM-powered agents from different sources outputting answers to the same query. The pairwise ranking is used to evaluate and aggregate their responses into a final solution.

    \item \textbf{ReAct.} Following~\citep{yao2023react}, this operator enables the agent to leverage versatile tools, including code interpreter, web searching, external knowledge database, \textit{etc.}, to handle diverse user demands. 

    \item \textbf{ExpertPrompt.}  
    Similar to AutoGPT~\citep{autogpt} and expert prompting~\citep{xu2023expertprompting}, this operator employs dynamic control flows to allow the agent to decide which expert should be utilized for the task.
\end{enumerate}

We respectfully note that the selection of these operators is highly customizable, allowing users the flexibility to incorporate their desired operators into the operator repository of \ourmethod.

\section{Prompt Repository}\label{app:prompt}

\subsection{Tag Generation Prompt}\label{app:prompt_tag}

\begin{tcolorbox}[notitle, sharp corners, breakable, colframe=Periwinkle, colback=white, 
       boxrule=3pt, boxsep=0.5pt, enhanced, 
       shadow={3pt}{-3pt}{0pt}{opacity=1,mygrey},
       title={Tag generation},]
       \footnotesize
       {\fontfamily{pcr}\selectfont
\begin{lstlisting}
TAG_PROMPT = """
**Workflow Information**
- **Name:** {NAME}
- **Description:** {DESCRIPTION}
- **Code:** {CODE}

**Your Task**
This workflow is designed to address specific problems in the MATH dataset, which contains challenging, competition-level mathematics problems. 
Please generate five relevant tags for this workflow, focusing on the academic disciplines involved and the workflow's level of complexity.

Here is the task this workflow has successfully solved:

{TASK}

**Your goal is to design tags for this workflow so that, when a similar task arises, the workflow's tags will have the highest cosine similarity score with the new task's tags.**

**Examples**
1. **Example 1**
   - **Code:** {MATH}
   - **Tags:** Right Triangle, Intermediate Combinatorics, Intermediate Computational Mathematics, Intermediate Permutations, Intermediate Geometry
   
2. **Example 2**
   - **Code:** {CUSTOM}
   - **Tags:** Number Theory, Integer Properties, Relatively Prime, Prime Factors, Simple Mathematical Problems

**Output Format**
- Provide **5 tags**, separated by commas.
- Tags should reflect the primary academic disciplines or difficulty level associated with the MATH dataset.

**Guidelines**
1. **Focus Areas:**
   - **Academic Disciplines:** Identify the main fields related to the MATH dataset (e.g., Linear Algebra, Mathematics, Calculus).
   - **Problem Difficulty:** Assess the complexity level of the problems (e.g., Beginner, Intermediate, Advanced).

2. **Formatting:**
   - Do not include any additional text or explanations.
   - Ensure the output is a single line containing exactly five tags.

**Important Notes**
- **Avoid General Tags:** Do not use overly broad tags such as Artificial Intelligence, Natural Language Processing, or Cognitive Science.
- **Relevance to MATH:** Ensure the tags are specifically relevant to the MATH dataset's focus on mathematical problems and their difficulty levels.

**WRONG Implementation Examples**
1. Artificial Intelligence, Natural Language Processing, Reasoning Systems, Advanced Problem Solving, Cognitive Science, Multi-Agent System, AI-enhanced Problem Solving, Problem Solving
   - *Issue:* These tags are too general and do not focus on the difficulty level or specific academic disciplines related to the MATH dataset.
"""
\end{lstlisting}
}
\end{tcolorbox}

\subsection{Offspring Generation Prompt}\label{app:prompt_generate}

\begin{tcolorbox}[notitle, sharp corners, breakable, colframe=ForestGreen, colback=white, 
       boxrule=3pt, boxsep=0.5pt, enhanced, 
       shadow={3pt}{-3pt}{0pt}{opacity=1,mygrey},
       title={Offspring generation},]
       \footnotesize
       {\fontfamily{pcr}\selectfont
\begin{lstlisting}
PROMPT = """
You are an expert machine learning researcher specializing in the design of agent-based workflows. Your goal is to optimize existing architectures and create a highly efficient, effective, and economically viable multi-agent workflow that solves a specific query from the MATH dataset, which contains challenging, competition-level mathematics problems.

Leverage your extensive knowledge of LLM prompting techniques and agent workflows from existing literature to analyze the provided architectures. Extract valuable insights and lessons, and draw inspiration from related LLM agent papers or research in other fields to design a novel, creative architecture. THINK OUTSIDE THE BOX.

## Query:
Your task is to develop an improved multi-agent workflow that surpasses all existing workflows in accurately and efficiently solving the following query. Consider the difficulty, complexity, and discipline of the query to structure an innovative multi-agent workflow best suited to solve it.
{QUERY}

## Reference Multi-Agent Workflow:
You have several multi-agent workflow designs to serve as references.
{PARENTS}

## Multi-Agent Communication Structure Design Instructions:

To improve the efficiency, effectiveness, and cost-effectiveness of communication within the Multi-Agent workflow, refer to the following communication structures when designing the workflow. Successful implementations typically do not rely on complex frameworks or specialized libraries. Instead, they emphasize building with simple, composable patterns. Below are several structures you can consider. You are also encouraged to use your imagination and logical thinking to design even more suitable structures for solving the specific task:

{STRUCTURES}

## Output Instruction:
{OUTPUT_INSTRUCTION}

Your response should be in JSON format, adhering to the structure demonstrated in the example below:
{EXAMPLE}

## Common Mistakes:
Here are some common mistakes you might make:
{WRONG_IMPLEMENTATION}
"""

\end{lstlisting}
}
\end{tcolorbox}

\subsection{Mutation Prompt}\label{app:prompt_muta}

\subsubsection{LLM Mutation}\label{app:change_llm}

\begin{tcolorbox}[notitle, sharp corners, breakable, colframe=YellowGreen, colback=white, 
       boxrule=3pt, boxsep=0.5pt, enhanced, 
       shadow={3pt}{-3pt}{0pt}{opacity=1,mygrey},
       title={LLM Mutation},]
       \footnotesize
       {\fontfamily{pcr}\selectfont
\begin{lstlisting}
"""
1. **Large Language Model Mutation**  
   You can replace the LLM backbone that initializes the operators. Your options are limited to the following 4 choices:
   - meta-llama/llama-3.1-70b-instruct
   - qwen/qwen-2.5-72b-instruct
   - deepseek/deepseek-chat-v2.5
   - nousresearch/hermes-3-llama-3.1-70b
"""
\end{lstlisting}
}
\end{tcolorbox}

\subsubsection{Prompt Mutation}\label{app:change_prompt}

\begin{tcolorbox}[notitle, sharp corners, breakable, colframe=YellowGreen, colback=white, 
       boxrule=3pt, boxsep=0.5pt, enhanced, 
       shadow={3pt}{-3pt}{0pt}{opacity=1,mygrey},
       title={Prompt Mutation},]
       \footnotesize
       {\fontfamily{pcr}\selectfont
\begin{lstlisting}
"""
2. **Prompt Mutation**  
   You can modify the prompts used by invoking nodes, such as incorporating few-shot examples or clarifying task instructions.  
   Prompt mutation can enhance the clarity of the agent's output. You can also create specific prompts to guide the operator in generating a logical response or facilitate communication between operators. 
   - Write your own prompt and use it in the Custom method within the workflow:
   ```python
   INSTRUCTION_PROMPT = '''Provide a comprehensive, step-by-step solution to the given mathematical problem. Utilize existing mathematical knowledge to solve the problem. Your response should include: 
       1. A clear restatement of the problem.
       2. An explanation of the mathematical concepts and theorems involved.
       3. A detailed, logical progression of steps leading to the solution.
       4. Clear explanations for each step, including the reasoning behind it.
       5. All mathematical expressions and equations in LaTeX format.
       6. Visual aids or diagrams if applicable (described in text).
       7. Make sure the final answer displayed in a boxed LaTeX format."
   response = await self.custom(input=task, instruction=INSTRUCTION_PROMPT)
   '''
   ```
   - You can also concatenate previously generated string results in the input to provide more comprehensive contextual information:
   ```python
   response = await self.custom(input=task + f"xxx:{{xxx}}, xxx:{{xxx}}", instruction=INSTRUCTION_PROMPT)
   ```

   The output from the Custom method can be placed anywhere in the workflow:
   ```python
   solution = await self.generate(problem=f"Here is the task: {{task}}, here is the response from other operators:{{response['response']}}")
   ```

   **Note**:
   - Avoid using single quotes in your code, as they may cause execution errors.
   - In the `custom` method, the input and instruction are directly concatenated (instruction + input), and placeholders are not supported. Be sure to handle concatenation externally and add comments where necessary.
"""
\end{lstlisting}
}
\end{tcolorbox}

\subsubsection{Operator Mutation}\label{app:change_topo}

% \subsubsection{Tool Mutation Agent}\label{app:change_tool}

\begin{tcolorbox}[notitle, sharp corners, breakable, colframe=YellowGreen, colback=white, 
       boxrule=3pt, boxsep=0.5pt, enhanced, 
       shadow={3pt}{-3pt}{0pt}{opacity=1,mygrey},
       title={Operator Mutation},]\label{box:tag-generate}
       \footnotesize
       {\fontfamily{pcr}\selectfont
\begin{lstlisting}
"""3. **Operator Mutation**  
   You can add or remove operators from the existing reference workflows. Consider their performance and compatibility with the given task.  
   Below are descriptions of the operators you can use. Initialize and call them properly, writing appropriate prompts to organize them and ensure they collaborate efficiently to solve the task.
   {OPERATORS}

   Additionally, remember to initialize operators in the `__init__` function before calling them!
"""

OPERATORS = {OPERATORS}
"""
\end{lstlisting}
}
\end{tcolorbox}

\section{History Management of \ourmethod}\label{app:history}

\subsection{LLM Experience Pool}\label{app:llm_pool}

To evaluate the historical performance of LLMs within agentic workflows, we construct an experience pool, denoted as \(\mathcal{P}_{LLM}\). This pool captures the interplay between LLM instances, prompts, and workflow configurations, providing a foundation for analyzing and refining their performance across diverse tasks.

\(\mathcal{P}_{LLM}\) captures both quantitative and qualitative evaluations of LLM's behavior across workflows. For a given workflow \(\mathcal{G}_k\) associated with task \(q\) and ground-truth answer \(a\), the performance of an LLM \(M_i\) is represented as \(\mathcal{F}_{LLM}(M_i, \mathcal{G}_k) = (\mathcal{R}_{LLM}, \mathcal{C}_{LLM})\), where \(\mathcal{R}_{LLM} \in \{\text{Positive}, \text{Negative}, \text{None}\}\) denotes a quantitative assessment of the LLM's output correctness. Specifically, \(\text{Positive}\) indicates that \(M_i\) produced a correct answer, \(\text{Negative}\) indicates an incorrect answer, and \(\text{None}\) signifies that \(M_i\) was not utilized in \(\mathcal{G}_k\). Additionally, \(\mathcal{C}_{LLM}\) provides a qualitative evaluation, offering detailed textual feedback on \(M_i\)'s role in the workflow, including how its behavior contributed to or detracted from solving the task. The overall experience pool is thus defined as \(\mathcal{P}_{LLM} = \bigcup_{i=1}^{|\mathcal{M}|} \{(M_i, \{(\mathcal{G}_k, \mathcal{R}_{LLM}, \mathcal{C}_{LLM}) \mid \forall \mathcal{G}_k \})\}\), aggregating performance data across all workflows and tasks. By capturing both the correctness and the nuanced role of each LLM in addressing diverse tasks, \(\mathcal{P}_{LLM}\) provides a comprehensive resource for understanding the strengths, weaknesses, and contextual suitability of different LLMs. This facilitates informed decision-making for LLM selection and adaptive workflow optimization.

\subsection{Workflow Experience Pool}\label{app:workflow_pool}
  
The workflow experience pool, denoted as \(\mathcal{P}_{WF}\), systematically captures the historical performance of workflows by maintaining a collection of records in the form of triplets \((\mathcal{G}_k, q, \mathcal{E}_{WF})\). Here, \(\mathcal{G}_k\) represents a specific workflow, \(\mathcal{Q}_j\) denotes a query or task associated with the workflow, and \(\mathcal{E}_{WF}\) is the corresponding evaluation of the workflow's performance on the given query. The evaluation \(\mathcal{E}_{WF} = (\mathcal{R}_{WF}, \mathcal{C}_{WF})\) consists of two components: a quantitative assessment \(\mathcal{R}_{WF} \in \{\text{Positive}, \text{Negative}\}\), which indicates whether the workflow successfully solved the query (\(\text{Positive}\)) or failed (\(\text{Negative}\)), and a qualitative assessment \(\mathcal{C}_{WF}\), which provides detailed textual feedback on the workflow's effectiveness, efficiency, and potential limitations in addressing the query. Formally, the experience pool is defined as \(\mathcal{P}_{WF} = \{(\mathcal{G}_k, \mathcal{Q}_j, \mathcal{E}_{WF}) \mid \forall \mathcal{G}_k, \mathcal{Q}_j\}\), aggregating evaluations across diverse workflows and queries. By systematically storing and analyzing these triplets, \(\mathcal{P}_{WF}\) offers a comprehensive resource for understanding the capabilities and limitations of various workflows, supporting iterative design refinements and enabling the development of more effective and adaptable agentic systems.

\section{Experimental Details}
\subsection{Dataset Statistics and Splits}\label{app:dataset}

Following existing practices in workflow automation~\cite{saad2024archon,hu2024adas,zhang2024aflow}, we partition each dataset with a \textsc{train:test} ratio of 1:4, except from ALFWorld dataset which follows the settings in \cite{shang2024agentsquare}. For the MATH benchmark, it is worth noting that we follow \cite{hong2024datainterpreter}, selecting 617 problems from four representative problem types (Combinatorics \& Probability, Number Theory, Pre-algebra, Pre-calculus) at difficulty level 5. The dataset statistics are concluded in \Cref{tab:dataset}.

\begin{table}[!h]
\vspace{-1em}
\caption{Dataset Statistics.}\label{tab:dataset}
% \begin{center}
\vspace{0.1em}
\centering
\begin{tabular}{l|cccc}
\toprule
Domain & Dataset & \#Train & \#Test & Metric  \\ 
\midrule
\multirow{2}{*}{Code Generation} & HumanEval & 33 & 131 & pass@1\\ 
& MBPP & 86 & 341 & pass@1\\ 
\midrule
\multirow{3}{*}{Math Reasoning}& GSM8K & 264 & 1055 & Accuracy\\ 
& MATH & 119 & 486 & Accuracy\\ 
& MultiArith &150 & 600 & Accuracy\\
\midrule
Embodied & ALFWorld & 230 & 327 & Success ratio\\
\bottomrule
\end{tabular}
\end{table}

\subsection{Baseline Setups}\label{app:baselines}

% \textbf{(1) manually designed workflows}, including Chain-of-Thought~\cite{cot}, ComplexCoT~\cite{fu2022complexity}), Self-Consistency (SC)~\cite{wang2023selfconsistency},  LLM-Debate~\citep{arXiv2023_MultiAgent-Debate}, LLM-Blender~\cite{blender}, DyLAN~\citep{arXiv2023_Dynamic-LLM-Agent}, AgentVerse~\citep{chen2023agentverse} and MacNet~\citep{qian2024scaling}; \textbf{(2) autonomous workflows}, including GPTSwarm~\citep{zhuge2024gptswarm}, AutoAgents~\citep{chen2023autoagents}, ADAS~\citep{hu2024adas}, AgentSquare~\citep{shang2024agentsquare} and AFlow~\citep{zhang2024aflow}

We detail the settings for all baselines in this section:  
\begin{enumerate}  
    \item \textbf{CoT.} CoT encourages LLM agents to reason step by step rather than directly producing an answer. We adopt the implementation from \cite{zhang2022automaticcot}.  

    \item \textbf{ComplexCoT.} The implementation is based on the code from \url{https://github.com/FranxYao/Complexity-Based-Prompting/tree/main}.  

    \item \textbf{Self-consistency.} We ensemble five CoT-generated solutions and adopt the implementation from \url{https://github.com/geekan/MetaGPT/blob/4954729e7564c806d7e58b3ed8b00ef991f889cc/metagpt/ext/aflow/scripts/operator.py#L93}.  

    \item \textbf{LLM-Debate.} We utilize five instances of the same LLM, assigning them distinct roles. These agents engage in up to two debate rounds, with the final answer determined via majority voting. Implementation follows \url{https://github.com/ucl-dark/llm_debate}.  

    \item \textbf{LLM-Blender.} The LLM-Blender is powered by two \llmname{gpt-4o-mini}, one \llmname{Qwen-2.5-72b}, and one \llmname{llama-3.1-70b}.  

    \item \textbf{DyLAN.} We directly adopt the implementation from \cite{arXiv2023_Dynamic-LLM-Agent}.  

    \item \textbf{AgentVerse.} The implementation is adopted from \cite{chen2023agentverse}.  

    \item \textbf{MacNet.} For MacNet~\citep{qian2024scaling}, we select the "MacNet-MESH" variant, which is essentially a densely connected complete graph.  

    \item \textbf{GPTSwarm.} We follow the original implementation and settings described in \cite{zhuge2024gptswarm}.  

    \item \textbf{AutoAgents.} The setup adheres to the original settings from \cite{chen2023autoagents}.  

    \item \textbf{ADAS.} Implementation details are directly adopted from \cite{hu2024adas}.  

    \item \textbf{AgentSquare.} We employ the modular search framework from \cite{shang2024agentsquare}. The base LLM is consistently set to \llmname{gpt-4o-mini}, with early stopping patience fixed at 5.  

    \item \textbf{AFlow.} In \cite{zhang2024aflow}, AFlow utilizes both \llmname{gpt-4o-mini} and the advanced \llmname{claude-3.5-sonnet}. To ensure fairness in homogeneous settings, we limit AFlow to \llmname{gpt-4o-mini} and set \textsc{max\_iteration}=20.  
\end{enumerate}  

\section{Supplementary Results}\label{app:exp-result}

\begin{table}[!h]
\centering
\caption{Performance comparison of different methods using various LLM backbones and training datasets. ‘’MATH'' and ``MBPP'' represent individual training datasets, while ``MATH+MBPP'' indicates training using both datasets combined. The two values under  ``MATH+MBPP'' represent the performance on MATH and MBPP, respectively.}
\label{tab:dylan_comparison}
\renewcommand{\arraystretch}{1.2}
\setlength{\tabcolsep}{8pt}

\begin{tabular}{l|l|ccc}
\Xhline{1.2pt}
\rowcolor{CadetBlue!20} 
\textbf{Method} & \textbf{LLM Backbone} & \textbf{MATH} & \textbf{MBPP} & \textbf{MATH+MBPP} \\ 
\Xhline{1.pt}
\multirow{2}{*}{DyLAN} & {\llmname{Deepseek-V2.5}} & $46.20$ & $80.13$ & $43.85/78.62$ \\ 
    & \llmname{QWen-2.5-72b} & $64.17$ & $75.63$ & $60.84/71.34$ \\ 
\midrule
\multirow{2}{*}{GPTSwarm} & \llmname{Deepseek-V2.5} & $45.36$ & $77.52$ & $39.18 / 74.09$   \\ 
& \llmname{QWen-2.5-72b} & $65.22$ & $72.48$ & $64.15 / 70.90$ \\ 
\midrule
\multirow{2}{*}{AFlow} & \llmname{Deepseek-V2.5} & $48.65$ & $79.14$ & $43.22 / 77.02$\\ 
 & \llmname{QWen-2.5-72b} & $66.38$ & $80.84$ & $64.71 / 74.90$ \\ 
\midrule
\ourmethod & \llmname{LLM Pool}  & $72.90$ & $87.62$ & $72.69 / 88.35$\\ 
                     
\Xhline{1.2pt}
\end{tabular}

\end{table}

%%%%%%%%%%%%%%%%%%%%%%%%%%%%%%%%%%%%%%%%%%%%%%%%%%%%%%%%%%%%%%%%%%%%%%%%%%%%%%%
%%%%%%%%%%%%%%%%%%%%%%%%%%%%%%%%%%%%%%%%%%%%%%%%%%%%%%%%%%%%%%%%%%%%%%%%%%%%%%%

\end{document}